\documentclass{article} 
\usepackage{iclr2027_conference,times}

\usepackage{amsmath,amsfonts,bm}

\def\eqref#1{equation~\ref{#1}}

\def\1{\bm{1}}

\DeclareMathAlphabet{\mathsfit}{\encodingdefault}{\sfdefault}{m}{sl}
\SetMathAlphabet{\mathsfit}{bold}{\encodingdefault}{\sfdefault}{bx}{n}

\usepackage{url}
\usepackage{amssymb}
\usepackage{graphicx}
\usepackage{capt-of}
\usepackage{booktabs}
\usepackage[table]{xcolor}
\usepackage{longtable}
\usepackage{arydshln}
\usepackage{pdflscape}
\usepackage{fvextra}
\usepackage{hyperref}

\title{\raggedright VISTA-Bench:\\
Benchmarking Multilingual Image\\
Translation with Image-Specific Rubrics}

\author{\makebox[\dimexpr\textwidth-2\tabcolsep\relax][c]{Bo Lv\thanks{Equal contribution.},\hspace{0.6em}Mao Zheng\footnotemark[1],\hspace{0.6em}Zheng Li\footnotemark[1],\hspace{0.6em}Fangxu Liu,\hspace{0.6em}Mingrui Sun,\hspace{0.6em}Tao Chen} \\
\multicolumn{1}{c}{Foundation Model Department, Tencent} \\
\multicolumn{1}{c}{\texttt{pokolv@tencent.com}}
}

\iclrfinalcopy
\begin{document}

\maketitle
\ificlrfinal
\lhead{VISTA-Bench}
\fancypagestyle{arxivfirstpage}{\lhead{\raisebox{-2pt}[8pt][0pt]{\includegraphics[height=18pt]{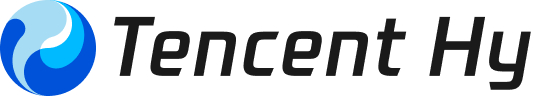}}}}
\thispagestyle{arxivfirstpage}
\fi

\begin{abstract}
Image translation is a fundamental capability of multimodal models for multilingual applications, requiring visual understanding and meaning preservation across languages.
However, existing benchmarks have limited language coverage and often lack explicit image-specific evaluation criteria, making it difficult to comprehensively assess this capability.
To systematically evaluate this capability, we introduce VISTA-Bench, covering 22 languages and 10 domains, and develop an image-specific rubric evaluation protocol.
The benchmark combines sampling for language and scenario coverage with model-assisted, human-verified annotations that group related text into coherent semantic units and provide multilingual reference translations.
The rubrics specify essential content, semantic relations, and acceptable translation variants, yielding separate output-based scores for translation quality and the preservation of visual and knowledge-dependent information.
We conduct extensive evaluations of 16 mainstream models, including 12 multimodal models and four text-input models, and provide systematic analyses across languages, domains, and evaluation dimensions.
The code and data will be available at \url{https://github.com/lvbotenbest/VISTA-BENCH}.
\end{abstract}

\section{Introduction}
\label{sec:introduction}

Image translation requires coordinating visual information acquisition, semantic interpretation, and target-language generation to preserve meaning across languages \citep{zhu-etal-2023-peit,lv-etal-2024-taekd}.
For multimodal large language models, this goes beyond recognizing isolated text: models must interpret relationships among text regions and use visual context to resolve ambiguity where necessary \citep{qian-etal-2024-anytrans}.
We study image-in, translated-text-out evaluation; text localization, layout reconstruction, background preservation, and translated-image quality are outside its scope.
A suitable benchmark must therefore combine diverse language and scene coverage with evaluation that accommodates valid translation variation while detecting omissions or mistranslations of essential information and semantic relationships.

Existing benchmarks have advanced visually situated translation, multilingual coverage, and position-aware evaluation \citep{salesky-etal-2024-benchmarking,li-etal-2025-mit,zhuang-etal-2025-patimt}.
These benchmarks employ automatic translation metrics such as BLEU \citep{papineni-etal-2002-bleu}, chrF \citep{popovic-2015-chrf}, and COMET \citep{rei-etal-2020-comet}.
Recent work also incorporates multimodal model-based judging \citep{Li_2026_CVPR}.
Nevertheless, source- and target-language coverage remains uneven, and individual benchmarks emphasize different subsets of real-world scenarios.
On the evaluation side, aggregate similarity scores and general quality judgments do not necessarily reveal whether a translation satisfies the specific requirements of an image.
Visual-dependence annotations and localization metrics provide useful complementary information, but they serve different purposes from explicitly specifying which meanings, relations, and acceptable interpretations should be preserved.

To address these needs, we introduce VISTA-Bench, comprising 2,228 images across 22 languages, 10 major domains, and 100+ fine-grained scenarios.
Its construction jointly considers source-language and scenario coverage, using sampling across language--scenario groups to obtain a compact evaluation set.
Model-assisted annotation and human verification establish the source text, semantic blocks, and multilingual reference translations (Appendix~\ref{app:annotation_team}).
In particular, annotators merge fragments that jointly express a semantic unit while retaining the separation of unrelated content.
This process organizes the benchmark around meaningful source content rather than treating individual text detections as independent translation units.

Prior work explores fine-grained MQM error annotation with LLMs \citep{kocmi-federmann-2023-gemba} and case-specific MQM rubrics \citep{xu2026rubricasexperts}.
Alongside the data, we develop an image-specific rubric evaluation protocol.
Each image receives a separate rubric covering visual understanding, translation quality, and knowledge-dependent interpretation where applicable.
Shared criteria are supplemented with target-language-specific requirements when needed.
The criteria specify required meaning, acceptable expressions, and disallowed errors, allowing legitimate variation while making content-specific constraints explicit.
A judge model assesses the candidate translation against these criteria and provides item-level decisions with supporting evidence.
Fixed rules aggregate the decisions by criterion importance within each dimension, yielding separate V, T, and K scores.
Our contribution thus combines coverage-oriented benchmark construction with an evaluation design that connects translation scores to identifiable requirements in individual images.

The remainder of this paper is organized as follows.
After reviewing related work, we describe the construction of VISTA-Bench and its image-specific rubric evaluation protocol.
We then benchmark 16 mainstream models, comprising 12 multimodal models and four text-input models, across 22 languages and 10 domains.
We analyze translation quality alongside visual understanding and knowledge use, examine variation across languages and domains, and assess dataset-size sensitivity and agreement between automatic judges and human annotators.

\section{Related Work}
\label{sec:related_work}

\subsection{Text-Image Machine Translation}
\label{sec:timt}

Text-image machine translation (TIMT) translates source-language text embedded in images into a target language \citep{lan-etal-2023-exploring}.
Existing approaches address both cross-modal alignment and contextual interpretation: PEIT uses pretrained models to bridge the vision--text modality gap in end-to-end translation \citep{zhu-etal-2023-peit}, while AnyTrans incorporates textual and visual context when translating fragmented text \citep{qian-etal-2024-anytrans}.
These complementary directions connect text perception with the preservation of meaning and relationships among text regions.

\subsection{Multilingual Image Translation Benchmarks}
\label{sec:multilingual_benchmarks}

Among existing multilingual image translation benchmarks \citep{salesky-etal-2024-benchmarking,li-etal-2025-mit,tian-etal-2025-prim,Li_2026_CVPR}, Vistra and PRIM both use only English source images,
with translations into four and five target languages, respectively.
Vistra focuses on visual context in natural images, whereas PRIM primarily covers images containing a single line of text
\citep{salesky-etal-2024-benchmarking,tian-etal-2025-prim}.
MIT-10M contains source images in eight languages, mainly drawn from web pages and product displays
\citep{li-etal-2025-mit}.
MMTIT-Bench includes images in fourteen source languages other than English and Chinese,
but provides translations only into Chinese and English, covering documents, web pages, and natural scenes
\citep{Li_2026_CVPR}.

Across these benchmarks, source--target coverage and scenario emphasis remain uneven.
Evaluation includes BLEU, chrF, and COMET \citep{salesky-etal-2024-benchmarking}, as well as multimodal judging with shared quality dimensions \citep{Li_2026_CVPR}.
In text translation, GEMBA-MQM identifies fine-grained error spans \citep{kocmi-federmann-2023-gemba}, while Rubric-as-Experts develops case-specific MQM rubrics \citep{xu2026rubricasexperts}.
Building on these lines of work, VISTA-Bench combines multilingual, multi-scenario coverage with image-specific rubrics that specify required meanings, semantic relations, and acceptable translation variants.

\section{Benchmark Construction and Evaluation}
\label{sec:benchmark_construction}

\begin{figure}[t]
\centering
\includegraphics[width=\linewidth]{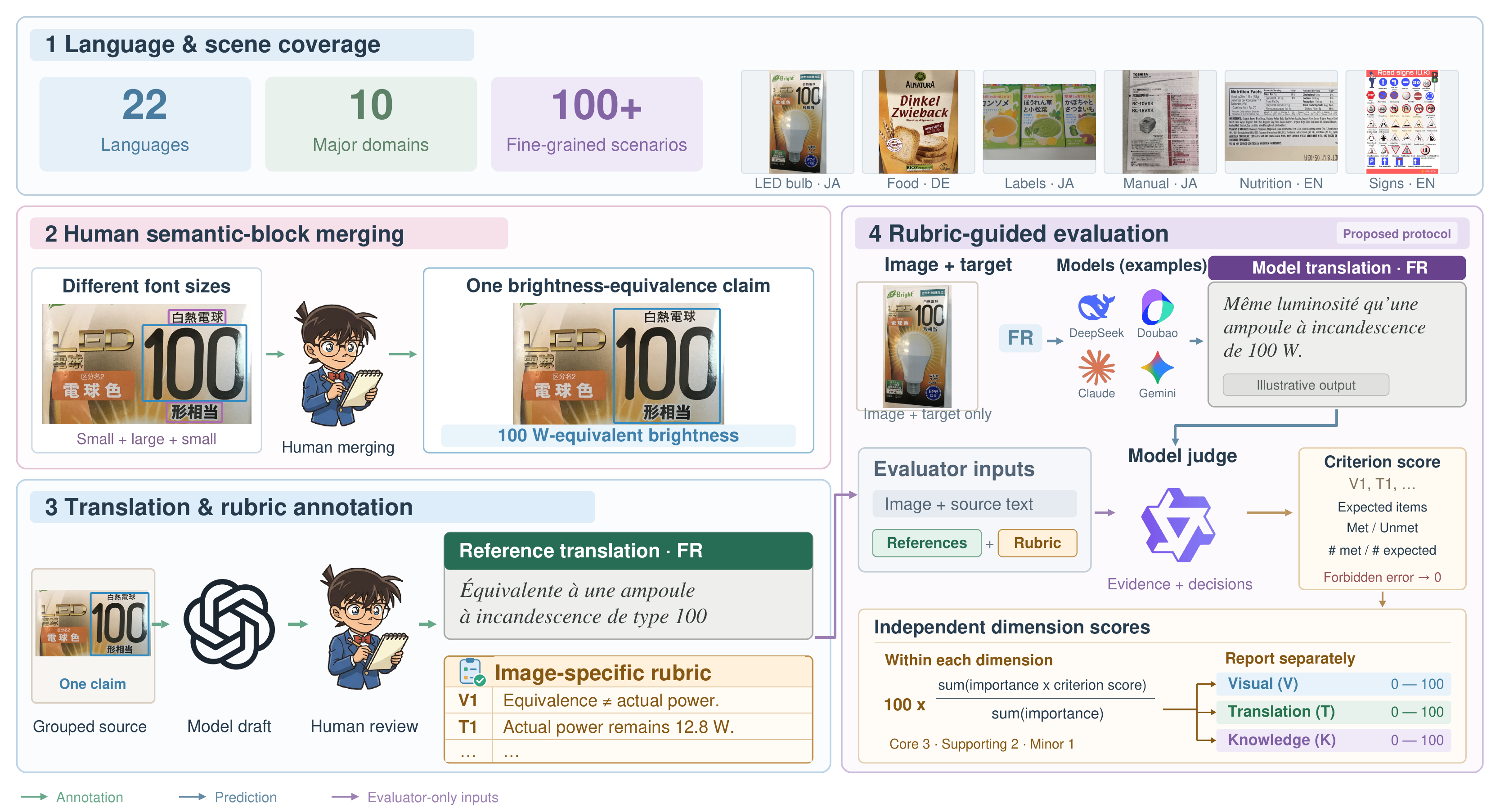}
\caption{Overview of VISTA-Bench. Stages 1--3 construct language--scenario coverage, human-reviewed semantic blocks, and multilingual references with image-specific rubrics. Stage 4 separates the model's image--target input from evaluator-only annotations, then converts evidence-based judgments into separate importance-weighted V, T, and K scores.}
\label{fig:vista_framework}
\end{figure}

VISTA-Bench comprises 2,228 images across 22 languages, 10 major domains, and 100+ fine-grained scenarios.
The annotation inventory distinguishes two Chinese scripts, giving 23 source/target codes and 49,016 reference slots; after merging script variants and excluding same-language tasks, there are 46,788 nominal image--target-language pairs.
Actual scoring coverage is smaller and dimension-specific.
Figure~\ref{fig:vista_framework} summarizes construction and evaluation.
Table~\ref{tab:benchmark_comparison} compares coverage and evaluation protocols across related benchmarks \citep{salesky-etal-2024-benchmarking,Li_2026_CVPR,tian-etal-2025-prim,tian-etal-2025-exploring,zhuang-etal-2025-patimt}; their text, rendered-image, and localization outputs are not identical.

\begin{table}[htbp]
\centering
\caption{Comparison of image translation benchmarks and evaluation protocols.}
\label{tab:benchmark_comparison}
\scriptsize
\setlength{\tabcolsep}{2.5pt}
\renewcommand{\arraystretch}{1.15}
\begin{tabular*}{\linewidth}{@{\extracolsep{\fill}}lrlll@{}}
\hline
\textbf{Dataset} & \textbf{Test Size} & \shortstack[l]{\textbf{Languages}\\(source $\rightarrow$ target)} & \textbf{\# Scenarios} & \textbf{Evaluation} \\
\hline
Vistra & 772 & En$\rightarrow$De/Es/Ru/Zh & No scenario breakdown & BLEU, chrF, COMET \\[2pt]
MMTIT-Bench & 1,400 & 14 languages $\rightarrow$ En/Zh & 3 (broad types) & COMET + VLLM judge \\[2pt]
PRIM & 340 & En$\rightarrow$De/Fr/Cs/Ru/Ro & No scenario breakdown & BLEU, COMET \\[2pt]
IIMT30K & 2,740 & En$\leftrightarrow$De & No scenario breakdown & BLEU, COMET \\[2pt]
PATIMT-Bench & 1,200 & En$\leftrightarrow$Zh & 10 (6 eval. groups) & BLEU, COMET \\[2pt]
\hline
VISTA-Bench (Ours) & 2,228 & \textbf{22 languages $\rightarrow$ 22 languages} & \textbf{100+ (10 domains)} & \textbf{Image-specific rubric judge} \\
\hline
\end{tabular*}
\end{table}

\subsection{Language and Scenario Coverage}
\label{sec:coverage_sampling}

Images from existing collections, local data, and web retrieval undergo source-specific automatic filtering, exact deduplication, and sensitive-information handling (Figure~\ref{fig:vista_framework}, stage 1).
We then assign language labels and classify images using a two-level taxonomy of major domains and fine-grained usage scenarios, producing a candidate pool of 10,920 images.
For example, printed menus are grouped under dining, road signs under transportation, and textbook content under publications.

To cover diverse use cases within the ten major domains, we use their fine-grained scenarios as sampling units.
This allows each domain to be represented by images from different subscenarios while keeping the evaluation set compact.
Let \(G_1,\ldots,G_K\) denote the nonempty groups, each containing images with the same source language and fine-grained scenario.
We form an initial subset by drawing a random sample from each group:
\begin{equation}
x_k\sim\operatorname{Uniform}(G_k),\quad k=1,\ldots,K,
\qquad \mathcal S_0=\{x_k\}_{k=1}^{K}.
\label{eq:stratified_random_sampling}
\end{equation}
We then expand this initial selection by sampling additional images from the fine-grained scenario groups in turn within each language, until the target sample size is reached.
All selected images undergo human review.

\subsection{Human Semantic-Block Merging}
\label{sec:semantic_grouping}

\noindent
\begin{minipage}[t]{0.49\linewidth}
\vspace{0pt}
\raggedright
Accurate image translation relies on understanding how text regions jointly convey meaning, rather than interpreting each fragment in isolation.
We use \texttt{gpt5.6 sol} for OCR and layout-aware semantic grouping (Figure~\ref{fig:vista_framework}, stage 2).
Annotators correct text, omissions, reading order, and block boundaries against the images, merging fragments that express a shared meaning while keeping unrelated content separate.

\smallskip
``Any revision'' denotes the proportion of images with at least one text or structural change, counting each image only once.
Across the ten languages in Figure~\ref{fig:human_revision_rates}, the mean per-language rate of text or structural revision is 42.7\%.
Rates are computed on images with comparable machine and human bounding boxes, excluding text-only annotations.
Structural revisions denote unmatched blocks under one-to-one matching at IoU $\geq 0.5$, excluding coordinate-only edits to matched blocks.
\end{minipage}\hfill
\begin{minipage}[t]{0.48\linewidth}
\vspace{0pt}
\centering
\setlength{\parskip}{0pt}
\setlength{\abovecaptionskip}{4pt}
\includegraphics[width=\linewidth]{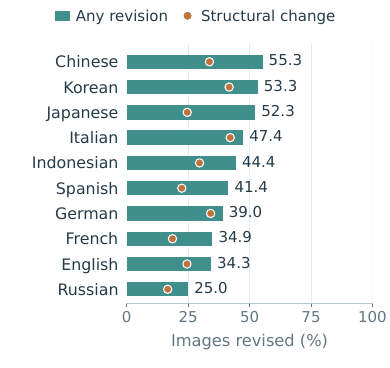}
\captionof{figure}{Human revision rates for ten selected widely used languages.}
\label{fig:human_revision_rates}
\end{minipage}
\par\medskip

\subsection{Reference and Rubric Annotation}
\label{sec:reference_rubric_annotation}

We use \texttt{gpt5.6 sol} to draft both reference translations and rubrics, with human verification at each stage (Figure~\ref{fig:vista_framework}, stage 3).
Translations into the target languages are drafted from the verified source text, with web search to resolve uncertain names and specialized terms.
Human annotators check and revise these drafts against the images.
The verified source blocks and references then guide rubric drafting, followed by human review and refinement.
Appendix~\ref{app:rubric_example} provides a complete rubric example, explains its use, and documents the drafting and judge prompt templates.

\par\medskip
\noindent\begin{minipage}{\linewidth}
\centering
\includegraphics[width=\linewidth]{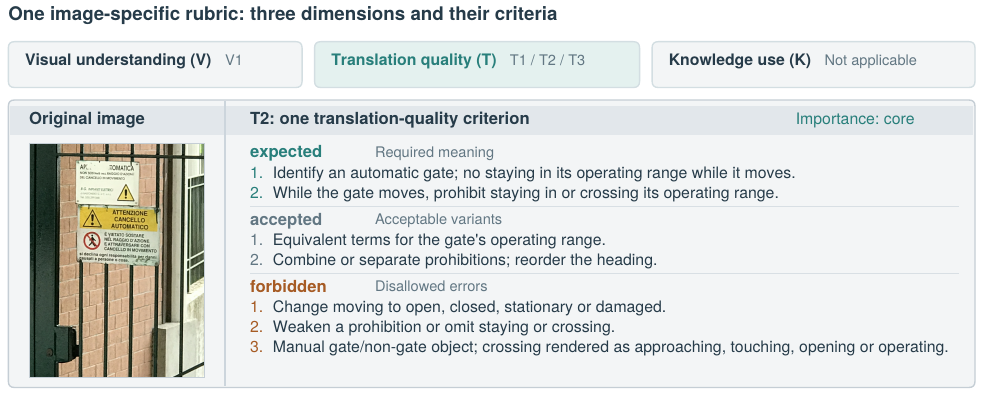}
\captionof{figure}{Rubric structure for an automatic-gate notice. Only T2 is expanded, with condensed wording. The knowledge dimension is inapplicable to this image.}
\label{fig:rubric_case}
\end{minipage}
\par\medskip

To account for lexical ambiguity and variation in valid translations, we construct a separate rubric for each image to complement its verified reference translations.
Each rubric defines three output-based diagnostics: \(V\) checks preservation of text coverage, grouping, and visual associations; \(T\) checks semantic fidelity and natural target-language expression; and \(K\) checks knowledge-dependent interpretations.
Knowledge criteria apply only where needed and do not reward additional facts.
These scores describe which requirements the translation satisfies, not whether the model actually relied on a particular visual cue or knowledge source.

For image \(i\) and target language \(\ell\), the evaluation criteria are
\begin{equation}
\mathcal R_{i,\ell}=\mathcal R_i\cup\Delta\mathcal R_{i,\ell},
\label{eq:rubric_adaptation}
\end{equation}
where \(\mathcal R_i\) contains the image's shared criteria and \(\Delta\mathcal R_{i,\ell}\) contains target-language-specific additions, with \(\Delta\mathcal R_{i,\ell}=\varnothing\) when none are needed.
Each criterion specifies required meaning (\texttt{expected}), acceptable expressions (\texttt{accepted}), and disallowed errors (\texttt{forbidden}).
Its importance label (core, supporting, or minor) determines its weight in score aggregation (Section~\ref{sec:evaluation_method}).

\subsection{Evaluation Protocol}
\label{sec:evaluation_method}

Once the rubrics are established, the evaluated model generates a free-form translation \(\hat y_{i\ell}=M(I_i,\ell)\) given only the image \(I_i\) and target language \(\ell\), without access to the annotations.
A judge model then assesses this output against the corresponding rubric and verified source blocks, returning item-level judgments and supporting evidence rather than an overall rating.
The protocol uses fixed rules to compute scores from these judgments, restricting score assignment while leaving semantic judgments subject to bias.

The judge marks each expected item as met only when its requirement is fully satisfied, and as unmet otherwise.
For a given image--target pair, let \(n_j>0\) denote the total number of expected items in criterion \(j\) and \(n_j^{\mathrm{met}}\) the number met.
The score is the proportion of items met unless a forbidden error occurs.
Writing \(f_j=1\) when any listed forbidden error is found and \(f_j=0\) otherwise, we obtain
\begin{equation}
q_j =
\begin{cases}
0, & f_j=1,\\[3pt]
\dfrac{n_j^{\mathrm{met}}}{n_j}, & f_j=0.
\end{cases}
\label{eq:criterion_score}
\end{equation}
Only expected items enter the denominator.
Accepted variants can satisfy these items but earn no additional credit.
A forbidden error sets only the corresponding criterion score to zero.
Otherwise, meeting two of three expected items yields \(2/3\).

For each image--target pair, we compute a separate importance-weighted score for each applicable dimension,
\begin{equation}
s_{i\ell,d}=100\frac{\sum_{j\in\mathcal J_{i\ell,d}} w_jq_j}
{\sum_{j\in\mathcal J_{i\ell,d}} w_j},
\qquad d\in\{V,T,K\},\quad \mathcal J_{i\ell,d}\ne\varnothing.
\label{eq:dimension_score}
\end{equation}
Here, \(\mathcal J_{i\ell,d}\) contains the applicable criteria in dimension \(d\), and \(s_{i\ell,d}\) is its score on a 0--100 scale.
The importance weights \(w_j=3,2,1\) correspond to core, supporting, and minor criteria, respectively, and apply only within a dimension.
A dimension with no applicable criteria is marked as not applicable rather than assigned zero.
We report V, T, and K separately without combining them into a composite score.
T is the primary translation measure, while V and K provide complementary diagnostics.

\section{Experiments}
\label{sec:experiments}

\subsection{Models}
\label{sec:experimental_models}

We report results for 16 models: 12 multimodal models and four text-input baselines.
The multimodal group includes gpt-6-astra\footnote{\raggedright\urlstyle{same}\url{https://developers.openai.com/api/docs/models/gpt-6-astra}}, doubao-seed-2-1-pro-260915, gemini-3.8-flash\footnote{\raggedright\urlstyle{same}\url{https://deepmind.google/models/model-cards/gemini-3-8-flash}}, qwen3.8-max-0902\footnote{\raggedright\urlstyle{same}\url{https://www.alibabacloud.com/help/en/model-studio/qwen3-8-max}}, muse-spark-1.3\footnote{\raggedright\urlstyle{same}\url{https://dev.meta.ai/}}, claude-sonnet-5\footnote{\raggedright\urlstyle{same}\url{https://www.anthropic.com/news/claude-sonnet-5}}, qwen3.8-27b-fp8\footnote{\raggedright\urlstyle{same}\url{https://huggingface.co/Qwen/Qwen3.8-27B-FP8}}, deepseek-flash, gpt-5.6-luna\footnote{\raggedright\urlstyle{same}\url{https://developers.openai.com/api/docs/models/gpt-5.6-luna}}, glm-5.3-flash\footnote{\raggedright\urlstyle{same}\url{https://huggingface.co/zai-org/GLM-5.3-Flash}}, glm-4.1v-9b, and gemma-4-e4b~\citep{gemmateam2026gemma4}.
These models receive the original image and a target-language instruction, following Section~\ref{sec:evaluation_method}.

The four text-input baselines reported in Table~\ref{tab:multimodal_domains} are aya-23-8B-text~\citep{aryabumi2024aya23}, gemma-3-4b-it-text~\citep{gemmateam2025gemma3}, HY-MT2-30B-A3B-text~\citep{zheng2026hymt2familyfastefficient}, and Qwen3-4B-text~\citep{qwen2025qwen3}.
The \texttt{-text} suffix identifies text-input runs; no results for other text-input configurations are claimed here.
For the reported text-input results, source text is organized using human-annotated layouts, and the models do not receive the original image.

\subsection{Evaluation Scores}
\label{sec:experimental_scores}

We report T, V, and K independently on a 0--100 scale, with T as the primary measure and V and K as complementary diagnostics.
For image-input models, retained criterion scores are importance-weighted within each dimension and direction, and global scores average directions equally.
\mbox{Domain T} applies the same procedure within each domain over common valid directions.
K includes only applicable criteria.
The Overall column in Table~\ref{tab:multimodal_domains} denotes global T, not a composite of the three dimensions.

\subsection{Domain-Level Performance}
\label{sec:domain_results}

Table~\ref{tab:multimodal_domains} reports global and domain-level T for image-input models, grouped by weight availability, and four text-input baselines.
Rows are ordered by global T from low to high within each group; tinted cells mark the best value in each group and column.

\begin{table}[htbp]
\centering
\caption{Translation-quality (T) results. Overall denotes global T, not the composite score. Global T averages directions for image-input models and tasks for text-input models. Tinted cells mark the highest score within each group and column.}
\label{tab:multimodal_domains}
\scriptsize
\setlength{\tabcolsep}{1.5pt}
\renewcommand{\arraystretch}{1.28}
\setlength{\dashlinedash}{2pt}
\setlength{\dashlinegap}{2pt}
\begin{tabular*}{\linewidth}{@{\extracolsep{\fill}}l*{11}{r}@{}}
\toprule
\textbf{Model} & \textbf{Overall} & \textbf{Transport} & \textbf{Dining} & \textbf{Products} & \textbf{Tourism} & \textbf{\shortstack{Health\\care}} & \textbf{Safety} & \textbf{\shortstack{Social\\Media}} & \textbf{\shortstack{Publi-\\cations}} & \textbf{Finance} & \textbf{\shortstack{Public\\Services}} \\
\midrule
\multicolumn{12}{l}{\textbf{Image-input: closed-weight models}} \\
GPT-5.6 Luna & 55.88 & 51.00 & 60.03 & 53.44 & 54.38 & 61.84 & 60.94 & 58.92 & 58.59 & 58.35 & 63.55 \\
Claude Sonnet 5 & 69.43 & 64.71 & 71.78 & 65.71 & 71.14 & 74.18 & 77.94 & 67.08 & 71.38 & 78.13 & 77.82 \\
Muse Spark 1.3 & 71.56 & 64.09 & 71.98 & 70.13 & 71.99 & 77.43 & 77.59 & 73.70 & 76.86 & 72.75 & 73.95 \\
Qwen3.8 Max & 72.67 & 66.00 & 71.94 & 72.57 & 72.89 & 76.20 & 81.07 & 74.67 & 75.57 & 76.43 & 75.40 \\
Gemini 3.8 Flash & 74.20 & 69.10 & 75.62 & 74.57 & 76.22 & \cellcolor[HTML]{EAF1FB}79.52 & 79.12 & 72.64 & 76.34 & 78.40 & 78.17 \\
Doubao Seed 2.1 Pro & 75.12 & 68.43 & 73.48 & 75.66 & 74.24 & 77.60 & 81.89 & 77.48 & 78.38 & \cellcolor[HTML]{EAF1FB}80.20 & 78.87 \\
GPT-6 Astra & \cellcolor[HTML]{EAF1FB}77.63 & \cellcolor[HTML]{EAF1FB}73.78 & \cellcolor[HTML]{EAF1FB}76.03 & \cellcolor[HTML]{EAF1FB}76.11 & \cellcolor[HTML]{EAF1FB}77.02 & 76.98 & \cellcolor[HTML]{EAF1FB}83.73 & \cellcolor[HTML]{EAF1FB}81.63 & \cellcolor[HTML]{EAF1FB}80.19 & 79.70 & \cellcolor[HTML]{EAF1FB}79.17 \\
\hdashline
\multicolumn{12}{l}{\textbf{Image-input: open-weight models}} \\
Gemma-4 E4B & 19.54 & 20.55 & 12.46 & 17.12 & 23.60 & 23.85 & 27.74 & 23.53 & 17.65 & 15.83 & 30.80 \\
GLM-4.1V 9B & 29.96 & 30.39 & 25.36 & 27.78 & 31.06 & 36.42 & 33.67 & 31.99 & 32.86 & 38.78 & 35.71 \\
GLM-5.3 Flash & 55.33 & 52.41 & 55.44 & 53.89 & 54.47 & 63.23 & 60.37 & 55.24 & 60.74 & 64.68 & 60.62 \\
DeepSeek Flash & 61.04 & 52.41 & \cellcolor[HTML]{F0EBFA}62.95 & 57.80 & \cellcolor[HTML]{F0EBFA}63.44 & \cellcolor[HTML]{F0EBFA}69.86 & \cellcolor[HTML]{F0EBFA}69.91 & 62.51 & 65.88 & \cellcolor[HTML]{F0EBFA}68.92 & 68.15 \\
Qwen3.8 27B & \cellcolor[HTML]{F0EBFA}61.11 & \cellcolor[HTML]{F0EBFA}56.79 & 55.63 & \cellcolor[HTML]{F0EBFA}60.18 & 63.22 & 68.80 & 67.59 & \cellcolor[HTML]{F0EBFA}63.18 & \cellcolor[HTML]{F0EBFA}67.32 & 65.60 & \cellcolor[HTML]{F0EBFA}68.52 \\
\hdashline
\multicolumn{12}{l}{\textbf{Text-input models}} \\
Aya-23 8B & 47.43 & 48.97 & 41.83 & 52.95 & 51.41 & 54.14 & 49.67 & 46.79 & 46.73 & 40.91 & 48.96 \\
Gemma-3 4B IT & 55.99 & 56.22 & 52.28 & 60.13 & 57.92 & 63.03 & 58.72 & 54.22 & 55.29 & 50.11 & 58.52 \\
Qwen3 4B & 64.34 & 65.44 & 54.00 & 68.55 & 66.27 & 72.38 & 67.10 & 61.73 & 65.56 & 59.56 & 67.60 \\
HY-MT2 30B-A3B & \cellcolor[HTML]{E9F5ED}72.82 & \cellcolor[HTML]{E9F5ED}72.75 & \cellcolor[HTML]{E9F5ED}71.05 & \cellcolor[HTML]{E9F5ED}78.02 & \cellcolor[HTML]{E9F5ED}75.65 & \cellcolor[HTML]{E9F5ED}81.67 & \cellcolor[HTML]{E9F5ED}75.42 & \cellcolor[HTML]{E9F5ED}68.12 & \cellcolor[HTML]{E9F5ED}75.22 & \cellcolor[HTML]{E9F5ED}64.56 & \cellcolor[HTML]{E9F5ED}76.15 \\
\bottomrule
\end{tabular*}
\end{table}

Among the image-input models, GPT-6 Astra achieves the highest global T (77.63) and ranks first in eight of the ten domains.
The exceptions are Healthcare, led by Gemini 3.8 Flash (79.52), and Finance, led by Doubao Seed 2.1 Pro (80.20).
Within the open-weight image-input group, Qwen3.8 27B has the highest global T (61.11), while DeepSeek Flash leads in Dining, Tourism, Healthcare, Safety, and Finance.
These changes in ranking show why domain-level scores complement the global average.
Appendix~\ref{app:target_domain_results} reports domain T by translation target language, averaging over source languages.

With human-organized source text, HY-MT2 30B-A3B achieves a global T of 72.82, compared with 77.63 for GPT-6 Astra.
It exceeds the best image-input scores in Products (78.02 vs.\ 76.11) and Healthcare (81.67 vs.\ 79.52), but trails in Finance (64.56 vs.\ 80.20).
The supplied semantic blocks let the text-input model bypass image recognition and layout recovery, whereas image-input models must recover both the text and its organization from pixels.
This contrast highlights text grouping and reading order as additional requirements of image translation.

\subsection{Language-Level Performance}
\label{sec:language_results}

Figure~\ref{fig:language_profiles} reports source-language T profiles for six representative models.

\par\smallskip
\noindent\begin{minipage}{\linewidth}
\centering
\setlength{\parskip}{0pt}
\setlength{\abovecaptionskip}{3pt}
\setlength{\belowcaptionskip}{0pt}
\includegraphics[width=\linewidth]{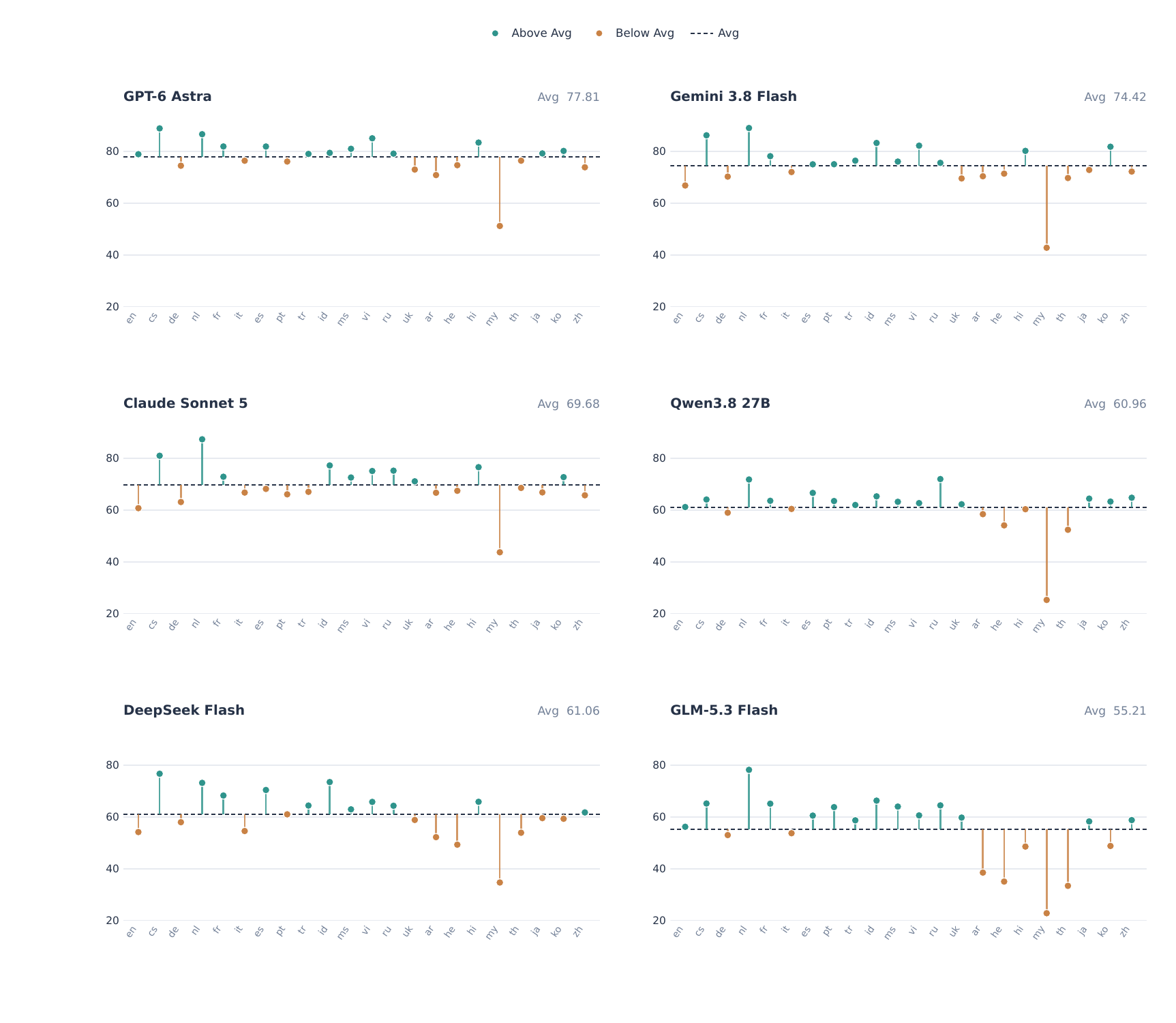}
\captionof{figure}{Source-language translation profiles for six multimodal models. Each point is the mean T for one source language over its target-language directions. Dashed lines mark the per-model average across 22 languages; teal and orange denote scores above and below that average.}
\label{fig:language_profiles}
\end{minipage}
\par

The profiles expose both common bottlenecks and model-specific strengths.
Burmese is the lowest-scoring source language for all six models, ranging from 22.90 for GLM-5.3 Flash to 51.22 for GPT-6 Astra.
Dutch is strongest for Gemini 3.8 Flash, Claude Sonnet 5, Qwen3.8 27B, and GLM-5.3 Flash, whereas Czech is strongest for GPT-6 Astra and DeepSeek Flash.
GPT-6 Astra obtains the highest score among these models on 19 of the 22 languages.
The three China-developed models shown all exceed their own language averages on Chinese: Qwen3.8 27B scores 64.79 versus 60.96, DeepSeek Flash 61.78 versus 61.06, and GLM-5.3 Flash 58.82 versus 55.21.
By contrast, GPT-6 Astra, Gemini 3.8 Flash, and Claude Sonnet 5 score 2.20--3.95 points below their respective averages on Chinese.
The shared difficulty on Burmese and differing language strengths show the value of broad source-language coverage: it exposes both common weaknesses and capabilities concentrated in particular languages.
These profiles identify where multilingual image translation needs further improvement.
Complete source-to-target T matrices for all 12 models are provided in Appendix~\ref{sec:full_direction_matrix}.

\subsection{Dimension-Level Analysis}
\label{sec:dimension_analysis}

Figure~\ref{fig:dimension_profiles} reports V and K alongside T for all 12 models, using the same T-based row order in both panels.
All three metrics use benchmark-wide direction averages; K retains its dimension-specific applicability.

\par\smallskip
\noindent\begin{minipage}{\linewidth}
\centering
\setlength{\parskip}{0pt}
\setlength{\abovecaptionskip}{3pt}
\setlength{\belowcaptionskip}{0pt}
\includegraphics[width=\linewidth]{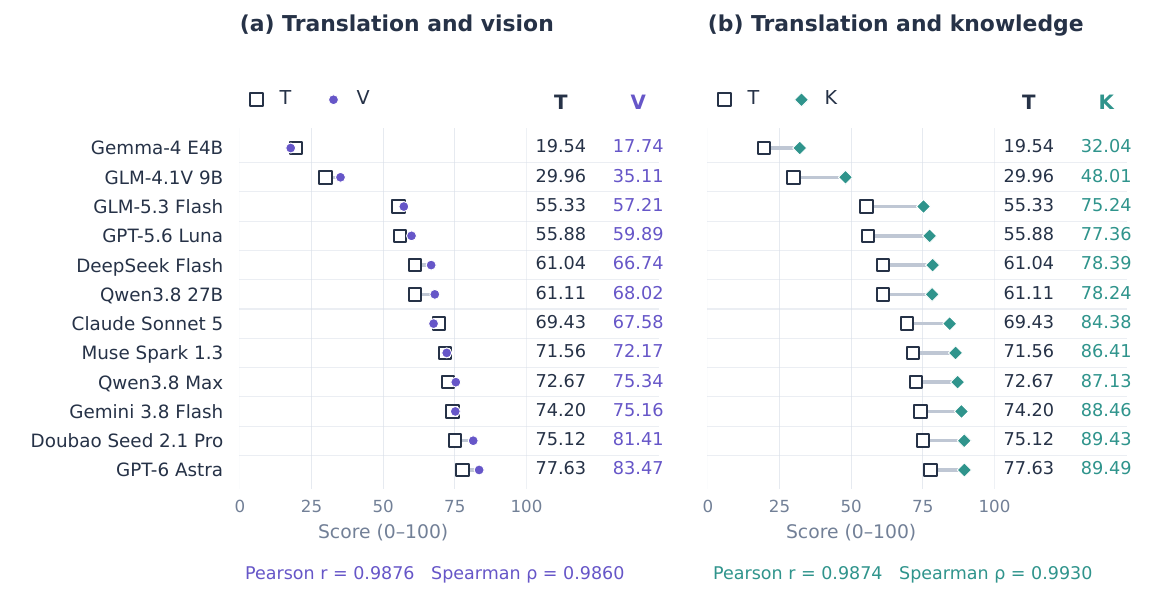}
\captionof{figure}{Paired T--V (left) and T--K (right) scores, ordered by increasing T from top to bottom. Lines connect the same model's scores; columns report values to two decimals. Correlations use all 12 models.}
\label{fig:dimension_profiles}
\end{minipage}
\par

GPT-6 Astra leads V and K (83.47 and 89.49), while Gemma-4 E4B scores 17.74 and 32.04.
T co-varies strongly with both diagnostics (Pearson \(r=0.9876\) for T--V and \(r=0.9874\) for T--K).
Despite similar V (67.58 vs.\ 68.02), Claude Sonnet 5 exceeds Qwen3.8 27B in K (84.38 vs.\ 78.24), alongside an 8.31-point advantage in T (69.43 vs.\ 61.11).
Appendix~\ref{app:source_dimension_results} provides T, K, and V for every model and source language.

\subsection{Dataset Size Ablation}
\label{sec:dataset_size_ablation}

We compare four models on the full pool of 10,920 images and the 2,228-image subset selected through language--scenario sampling (Section~\ref{sec:coverage_sampling}).
Table~\ref{tab:dataset_size_ablation} reports their translation-quality scores (T).

\par
\noindent\begin{minipage}{\linewidth}
\centering
\setlength{\parskip}{0pt}
\setlength{\abovecaptionskip}{3pt}
\captionof{table}{Translation-quality scores on the full and selected sets. Statistics are computed from the reported one-decimal scores.}
\label{tab:dataset_size_ablation}
\small
\setlength{\tabcolsep}{5pt}
\renewcommand{\arraystretch}{1.12}
\begin{tabular*}{0.94\linewidth}{@{\extracolsep{\fill}}lrrr@{}}
\toprule
\textbf{Model} & \shortstack{\textbf{Full set}\\(10,920)} & \shortstack{\textbf{Selected set}\\(2,228)} & \(\lvert\Delta T\rvert\) \\
\midrule
GLM-4.1V 9B & 30.7 & 30.0 & 0.7 \\
Gemma-4 E4B & 20.0 & 19.5 & 0.5 \\
Claude Sonnet 5 & 70.2 & 69.4 & 0.8 \\
GPT-6 Astra & 79.1 & 77.6 & 1.5 \\
\midrule
\multicolumn{4}{c}{Pearson \(r=0.99996\)\quad Spearman \(\rho=1.0000\)\quad MAE \(=0.88\)} \\
\bottomrule
\end{tabular*}
\end{minipage}
\par

The selected set retains about one fifth of the full pool while preserving the ranking of all four models.
T scores decrease by just 0.5--1.5 points, with an MAE of 0.88, showing close agreement between the smaller and larger evaluation sets.
The 2,228-image subset therefore provides a consistent basis for model comparison.
Using fewer images reduces the number of translation and judging requests, lowering the computational workload of evaluation and making repeated model comparisons more practical.
The selection spans source-language and fine-grained scenario groups, so the subset retains diverse language and scenario combinations rather than concentrating on a few frequent categories.
These findings support our selection method for building a diverse benchmark that reduces evaluation workload while retaining agreement with the full pool.

\subsection{Judge Robustness and Human Agreement}
\label{sec:judge_agreement}

We assess judge sensitivity on 440 images, sampling two images from each source-language--domain combination.
Translations from four candidate models into the selected target languages are evaluated by Qwen3.6, Qwen3.8 Max, and human annotators using the same image-specific rubrics.
For each automatic judge, scores are averaged over three judging runs.
Three human annotators make binary item-level judgments; majority decisions are converted to T scores using the same scoring rules.

\par
\noindent\begin{minipage}{\linewidth}
\centering
\setlength{\parskip}{0pt}
\setlength{\abovecaptionskip}{3pt}
\setlength{\belowcaptionskip}{0pt}
\includegraphics[width=\linewidth,trim=0bp 7bp 0bp 12bp,clip]{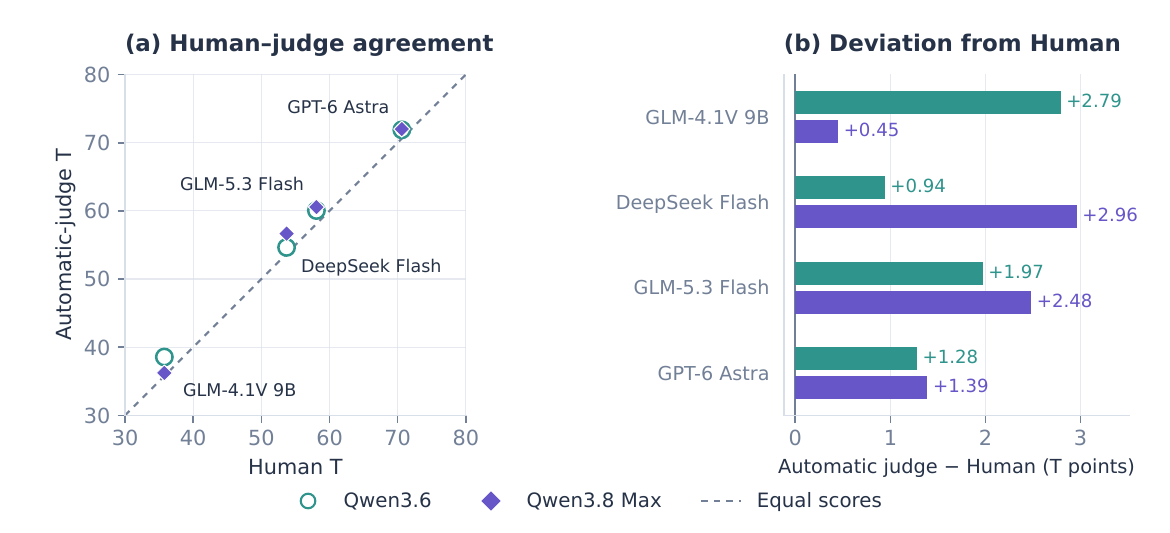}
\captionof{figure}{Human--judge agreement and score deviations. Left: each point is one candidate model's mean T under an automatic judge versus Human; the dashed line marks equal scores. Right: automatic-minus-human differences. Correlations use four model-level means, not individual images.}
\label{fig:judge_agreement}
\end{minipage}
\par

Both automatic judges preserve the human ordering of all four candidates (Spearman \(\rho=1.0000\); Figure~\ref{fig:judge_agreement}).
For Qwen3.6 and Qwen3.8 Max, respectively, Pearson correlations with Human are 0.9992 and 0.9976, and Lin's concordance correlation coefficients are 0.9883 and 0.9869.
The corresponding MAEs are 1.75 and 1.82 points.
All eight automatic scores are higher than their human references, indicating a small but consistent positive offset rather than exact score equivalence.
Switching between the two automatic judges changes model scores by 1.25 points on average and at most 2.34 points, without changing the ranking.

\section{Conclusion}
\label{sec:conclusion}

We introduced VISTA-Bench, comprising 2,228 images across 22 languages and 10 domains, with coverage-oriented sampling, human-verified semantic blocks, and multilingual references.
Our image-specific rubrics specify required meanings, semantic relations, and acceptable variants, yielding separate scores for translation quality, visual understanding, and knowledge use.
We evaluated 12 multimodal models and four text-input baselines.
The image-input results expose language and domain disparities and show that higher translation scores generally accompany stronger visual and knowledge scores.
The compact subset preserves the tested models' ranking while reducing evaluation workload.
Rubric-guided judges closely match human model-level assessments, with Qwen3.6 achieving agreement comparable to Qwen3.8 Max.
VISTA-Bench thus combines broad coverage with content-specific evaluation to support model comparison and targeted diagnosis.

\clearpage
\subsection*{AI use statement}

We used generative AI tools for language polishing and translation of the manuscript.
As described in Sections~\ref{sec:benchmark_construction} and~\ref{sec:judge_agreement}, generative models also assisted source-text recognition and semantic grouping, drafted multilingual reference translations and image-specific rubrics, and performed rubric-based evaluation.
Human annotators verified and revised the annotation drafts.
The authors reviewed all AI-assisted content adopted in this work and take responsibility for the final text, claims, and artifacts.

\subsection*{Ethics statement}

The images used in this study were obtained from publicly accessible sources and confirmed to be available for research use.
During data preparation, we excluded images containing sensitive personal information or de-identified the affected regions to reduce privacy risks.
Data sharing for this submission and any future public release will follow the applicable source licenses and permissions.
We will share image files only where redistribution is permitted; otherwise, we will provide source references and any annotations that we are permitted to release.

\subsection*{Reproducibility statement}

Section~\ref{sec:benchmark_construction} describes the benchmark construction and scoring procedures, and Section~\ref{sec:experiments} presents the evaluated models and experimental analyses.
Appendix~\ref{app:annotation_team} documents annotation quality control, Appendix~\ref{app:rubric_example} provides a complete rubric example and prompt templates, and Appendix~\ref{app:supplementary_experiments} reports detailed experimental results.
We will provide the evaluation code, configurations, and benchmark annotations as supplementary materials with the submission.
Data sharing will follow the source-license and permission constraints stated in the Ethics statement.

\bibliography{iclr2027_conference}

\begin{thebibliography}{20}
\providecommand{\natexlab}[1]{#1}
\providecommand{\url}[1]{\texttt{#1}}
\expandafter\ifx\csname urlstyle\endcsname\relax
  \providecommand{\doi}[1]{doi: #1}\else
  \providecommand{\doi}{doi: \begingroup \urlstyle{rm}\Url}\fi

\bibitem[Aryabumi et~al.(2024)Aryabumi, Dang, Talupuru, Dash, Cairuz, Lin,
  Venkitesh, Smith, Campos, Tan, Marchisio, Bartolo, Ruder, Locatelli,
  Kreutzer, Frosst, Gomez, Blunsom, Fadaee, {\"U}st{\"u}n, and
  Hooker]{aryabumi2024aya23}
Viraat Aryabumi, John Dang, Dwarak Talupuru, Saurabh Dash, David Cairuz, Hangyu
  Lin, Bharat Venkitesh, Madeline Smith, Jon~Ander Campos, Yi~Chern Tan, Kelly
  Marchisio, Max Bartolo, Sebastian Ruder, Acyr Locatelli, Julia Kreutzer, Nick
  Frosst, Aidan Gomez, Phil Blunsom, Marzieh Fadaee, Ahmet {\"U}st{\"u}n, and
  Sara Hooker.
\newblock {Aya 23}: Open weight releases to further multilingual progress.
\newblock \emph{arXiv preprint arXiv:2405.15032}, 2024.
\newblock \doi{10.48550/arXiv.2405.15032}.
\newblock URL \url{https://arxiv.org/abs/2405.15032}.

\bibitem[{Gemma Team}(2025)]{gemmateam2025gemma3}
{Gemma Team}.
\newblock {Gemma 3} technical report.
\newblock \emph{arXiv preprint arXiv:2503.19786}, 2025.
\newblock \doi{10.48550/arXiv.2503.19786}.
\newblock URL \url{https://arxiv.org/abs/2503.19786}.

\bibitem[{Gemma Team}(2026)]{gemmateam2026gemma4}
{Gemma Team}.
\newblock {Gemma 4} technical report.
\newblock \emph{arXiv preprint arXiv:2607.02770}, 2026.
\newblock \doi{10.48550/arXiv.2607.02770}.
\newblock URL \url{https://arxiv.org/abs/2607.02770}.

\bibitem[Kocmi \& Federmann(2023)Kocmi and
  Federmann]{kocmi-federmann-2023-gemba}
Tom Kocmi and Christian Federmann.
\newblock {GEMBA-MQM}: Detecting translation quality error spans with {GPT-4}.
\newblock In \emph{Proceedings of the Eighth Conference on Machine
  Translation}, pp.\  768--775. Association for Computational Linguistics,
  2023.
\newblock \doi{10.18653/v1/2023.wmt-1.64}.
\newblock URL \url{https://aclanthology.org/2023.wmt-1.64/}.

\bibitem[Lan et~al.(2023)Lan, Yu, Li, Zhang, Luan, Wang, Huang, and
  Su]{lan-etal-2023-exploring}
Zhibin Lan, Jiawei Yu, Xiang Li, Wen Zhang, Jian Luan, Bin Wang, Degen Huang,
  and Jinsong Su.
\newblock Exploring better text image translation with multimodal codebook.
\newblock In \emph{Proceedings of the 61st Annual Meeting of the Association
  for Computational Linguistics (Volume 1: Long Papers)}, pp.\  3479--3491.
  Association for Computational Linguistics, 2023.
\newblock \doi{10.18653/v1/2023.acl-long.192}.
\newblock URL \url{https://aclanthology.org/2023.acl-long.192/}.

\bibitem[Li et~al.(2025)Li, Zhu, and Wen]{li-etal-2025-mit}
Bo~Li, Shaolin Zhu, and Lijie Wen.
\newblock {MIT-10M}: A large scale parallel corpus of multilingual image
  translation.
\newblock In \emph{Proceedings of the 31st International Conference on
  Computational Linguistics}, pp.\  5154--5167. Association for Computational
  Linguistics, 2025.
\newblock URL \url{https://aclanthology.org/2025.coling-main.346/}.

\bibitem[Li et~al.(2026)Li, Zhang, Liang, Shen, Zhang, Lyu, Wang, Wan, Zeng,
  Hu, Ma, and Zhou]{Li_2026_CVPR}
Gengluo Li, Chengquan Zhang, Yupu Liang, Huawen Shen, Yaping Zhang, Pengyuan
  Lyu, Weinong Wang, Xingyu Wan, Gangyan Zeng, Han Hu, Can Ma, and Yu~Zhou.
\newblock {MMTIT-Bench}: A multilingual and multi-scenario benchmark with
  cognition-perception-reasoning guided text-image machine translation.
\newblock In \emph{Proceedings of the IEEE/CVF Conference on Computer Vision
  and Pattern Recognition (CVPR)}, pp.\  16593--16602, June 2026.

\bibitem[Lv et~al.(2024)Lv, Liu, Wei, Luo, and Yu]{lv-etal-2024-taekd}
Bo~Lv, Xin Liu, Kaiwen Wei, Ping Luo, and Yue Yu.
\newblock {TA}e{KD}: Teacher assistant enhanced knowledge distillation for
  closed-source multilingual neural machine translation.
\newblock In Nicoletta Calzolari, Min-Yen Kan, Veronique Hoste, Alessandro
  Lenci, Sakriani Sakti, and Nianwen Xue (eds.), \emph{Proceedings of the 2024
  Joint International Conference on Computational Linguistics, Language
  Resources and Evaluation (LREC-COLING 2024)}, pp.\  15530--15541, Torino,
  Italia, May 2024. ELRA and ICCL.
\newblock URL \url{https://aclanthology.org/2024.lrec-main.1350/}.

\bibitem[Papineni et~al.(2002)Papineni, Roukos, Ward, and
  Zhu]{papineni-etal-2002-bleu}
Kishore Papineni, Salim Roukos, Todd Ward, and Wei-Jing Zhu.
\newblock {Bleu}: a method for automatic evaluation of machine translation.
\newblock In \emph{Proceedings of the 40th Annual Meeting of the Association
  for Computational Linguistics}, pp.\  311--318. Association for Computational
  Linguistics, 2002.
\newblock \doi{10.3115/1073083.1073135}.
\newblock URL \url{https://aclanthology.org/P02-1040/}.

\bibitem[Popovi\'{c}(2015)]{popovic-2015-chrf}
Maja Popovi\'{c}.
\newblock chr{F}: character n-gram {F}-score for automatic {MT} evaluation.
\newblock In \emph{Proceedings of the Tenth Workshop on Statistical Machine
  Translation}, pp.\  392--395. Association for Computational Linguistics,
  2015.
\newblock \doi{10.18653/v1/W15-3049}.
\newblock URL \url{https://aclanthology.org/W15-3049/}.

\bibitem[Qian et~al.(2024)Qian, Zhang, Yang, Fan, Ma, Wong, Sun, and
  Ji]{qian-etal-2024-anytrans}
Zhipeng Qian, Pei Zhang, Baosong Yang, Kai Fan, Yiwei Ma, Derek~F. Wong,
  Xiaoshuai Sun, and Rongrong Ji.
\newblock {AnyTrans}: Translate {AnyText} in the image with large scale models.
\newblock In \emph{Findings of the Association for Computational Linguistics:
  EMNLP 2024}, pp.\  2432--2444. Association for Computational Linguistics,
  2024.
\newblock \doi{10.18653/v1/2024.findings-emnlp.137}.
\newblock URL \url{https://aclanthology.org/2024.findings-emnlp.137/}.

\bibitem[{Qwen Team}(2025)]{qwen2025qwen3}
{Qwen Team}.
\newblock {Qwen3} technical report.
\newblock \emph{arXiv preprint arXiv:2505.09388}, 2025.
\newblock \doi{10.48550/arXiv.2505.09388}.
\newblock URL \url{https://arxiv.org/abs/2505.09388}.

\bibitem[Rei et~al.(2020)Rei, Stewart, Farinha, and Lavie]{rei-etal-2020-comet}
Ricardo Rei, Craig Stewart, Ana~C Farinha, and Alon Lavie.
\newblock {COMET}: A neural framework for {MT} evaluation.
\newblock In \emph{Proceedings of the 2020 Conference on Empirical Methods in
  Natural Language Processing (EMNLP)}, pp.\  2685--2702. Association for
  Computational Linguistics, 2020.
\newblock \doi{10.18653/v1/2020.emnlp-main.213}.
\newblock URL \url{https://aclanthology.org/2020.emnlp-main.213/}.

\bibitem[Salesky et~al.(2024)Salesky, Koehn, and
  Post]{salesky-etal-2024-benchmarking}
Elizabeth Salesky, Philipp Koehn, and Matt Post.
\newblock Benchmarking visually-situated translation of text in natural images.
\newblock In \emph{Proceedings of the Ninth Conference on Machine Translation},
  pp.\  1167--1182. Association for Computational Linguistics, 2024.
\newblock \doi{10.18653/v1/2024.wmt-1.115}.
\newblock URL \url{https://aclanthology.org/2024.wmt-1.115/}.

\bibitem[Tian et~al.(2025{\natexlab{a}})Tian, Liu, Liu, Feng, Li, Huang, and
  Guo]{tian-etal-2025-prim}
Yanzhi Tian, Zeming Liu, Zhengyang Liu, Chong Feng, Xin Li, Heyan Huang, and
  Yuhang Guo.
\newblock {PRIM}: Towards practical in-image multilingual machine translation.
\newblock In \emph{Proceedings of the 2025 Conference on Empirical Methods in
  Natural Language Processing}, pp.\  13682--13697. Association for
  Computational Linguistics, 2025{\natexlab{a}}.
\newblock \doi{10.18653/v1/2025.emnlp-main.691}.
\newblock URL \url{https://aclanthology.org/2025.emnlp-main.691/}.

\bibitem[Tian et~al.(2025{\natexlab{b}})Tian, Liu, Liu, and
  Guo]{tian-etal-2025-exploring}
Yanzhi Tian, Zeming Liu, Zhengyang Liu, and Yuhang Guo.
\newblock Exploring in-image machine translation with real-world background.
\newblock In \emph{Findings of the Association for Computational Linguistics:
  ACL 2025}, pp.\  124--137. Association for Computational Linguistics,
  2025{\natexlab{b}}.
\newblock \doi{10.18653/v1/2025.findings-acl.6}.
\newblock URL \url{https://aclanthology.org/2025.findings-acl.6/}.

\bibitem[Xu et~al.(2026)Xu, Shen, Wang, Dang, and Huang]{xu2026rubricasexperts}
Weilu Xu, Yunzhi Shen, Xinye Wang, Ranfei Dang, and Shujian Huang.
\newblock {Rubric-as-Experts}: Case-specific {MQM} rubrics for translation
  quality evaluation.
\newblock \emph{arXiv preprint arXiv:2606.21559}, 2026.
\newblock \doi{10.48550/arXiv.2606.21559}.
\newblock URL \url{https://arxiv.org/abs/2606.21559v1}.

\bibitem[Zheng et~al.(2026)Zheng, Li, Chen, Lv, Sun, Song, Song, Huang, Wu,
  Wang, Song, Chen, and Zhang]{zheng2026hymt2familyfastefficient}
Mao Zheng, Zheng Li, Tao Chen, Bo~Lv, Mingrui Sun, Mingyang Song, Jinlong Song,
  Hong Huang, Decheng Wu, Hai Wang, Yifan Song, Yanfeng Chen, and Guanwei
  Zhang.
\newblock Hy-mt2: A family of fast, efficient and powerful multilingual
  translation models in the wild, 2026.
\newblock URL \url{https://arxiv.org/abs/2605.22064}.

\bibitem[Zhu et~al.(2023)Zhu, Li, Lei, and Xiong]{zhu-etal-2023-peit}
Shaolin Zhu, Shangjie Li, Yikun Lei, and Deyi Xiong.
\newblock {PEIT}: Bridging the modality gap with pre-trained models for
  end-to-end image translation.
\newblock In \emph{Proceedings of the 61st Annual Meeting of the Association
  for Computational Linguistics (Volume 1: Long Papers)}, pp.\  13433--13447.
  Association for Computational Linguistics, 2023.
\newblock \doi{10.18653/v1/2023.acl-long.751}.
\newblock URL \url{https://aclanthology.org/2023.acl-long.751/}.

\bibitem[Zhuang et~al.(2025)Zhuang, Li, Lan, Han, Li, and
  Su]{zhuang-etal-2025-patimt}
Wanru Zhuang, Wenbo Li, Zhibin Lan, Xu~Han, Peng Li, and Jinsong Su.
\newblock {PATIMT-Bench}: A multi-scenario benchmark for position-aware text
  image machine translation in large vision-language models.
\newblock In \emph{Findings of the Association for Computational Linguistics:
  EMNLP 2025}, pp.\  16572--16588. Association for Computational Linguistics,
  2025.
\newblock \doi{10.18653/v1/2025.findings-emnlp.900}.
\newblock URL \url{https://aclanthology.org/2025.findings-emnlp.900/}.

\end{thebibliography}
\bibliographystyle{iclr2027_conference}

\clearpage
\appendix
\renewcommand{\thetable}{\thesection\arabic{table}}
\renewcommand{\theHtable}{appendix.\thesection.\arabic{table}}
\setcounter{table}{0}
\raggedbottom
\section{Annotation Team and Quality Control}
\label{app:annotation_team}

Human annotation and verification involved nine team members, including two responsible for quality control.
The team included annotators with bachelor's or master's degrees in linguistics.
Their work covered source-text and semantic-block annotations, multilingual reference translations, and image-specific rubrics, as described in Sections~\ref{sec:semantic_grouping} and~\ref{sec:reference_rubric_annotation}.

For some languages, reviewers with direct proficiency in the relevant language could not be recruited.
In these cases, reviewers familiar with related languages used dictionaries and AI-assisted recognition to assess the annotations.
This assisted procedure is not equivalent to independent review by a fluent target-language specialist and does not establish uniform annotation quality across languages.

After initial annotation, the annotations underwent two rounds of verification combining cross-checks among annotators and review by the quality-control members.
Reviewers checked textual accuracy and completeness, semantic grouping and reading order, reference-translation fidelity, and the consistency of rubric requirements with the source image.
Potential omissions or inconsistencies identified during review were revisited against the source material, and the corresponding annotations were corrected.
Language-specific proficiency assignments, independent double-annotation rates, and inter-annotator agreement estimates are not reported.

\subsection{Source-Language Distribution}
\label{app:language_distribution}

Figure~\ref{fig:source_language_counts} shows the source-language composition of VISTA-Bench.

\par\smallskip
\noindent\begin{minipage}{\linewidth}
\centering
\includegraphics[width=\linewidth]{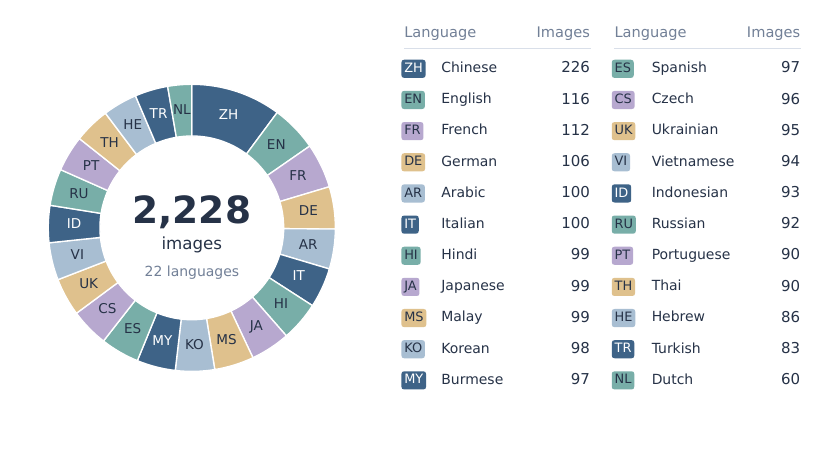}
\captionof{figure}{Source-image counts across 22 languages. Each of the 2,228 images is counted once; simplified and traditional Chinese are combined. Sector areas are proportional to image counts, and the legend reports absolute counts.}
\label{fig:source_language_counts}
\end{minipage}

\clearpage
\section{Complete Rubric Example and Prompt Templates}
\label{app:rubric_example}

\subsection{Example and Interpretation}
\label{app:rubric_example_context}

We use an English medicine-package label to illustrate a complete rubric for translation into French.
The label contains six source blocks, including the product name, quantity, ingredient, manufacturer, location, and logo identifier.
It is short enough to inspect in full while requiring all three evaluation dimensions: recovering the label's grouping, translating its content, and interpreting a place name consistently with the supplied knowledge annotation.
The example concerns translation of packaging text, not medical advice.

\begin{center}
\includegraphics[width=0.84\linewidth]{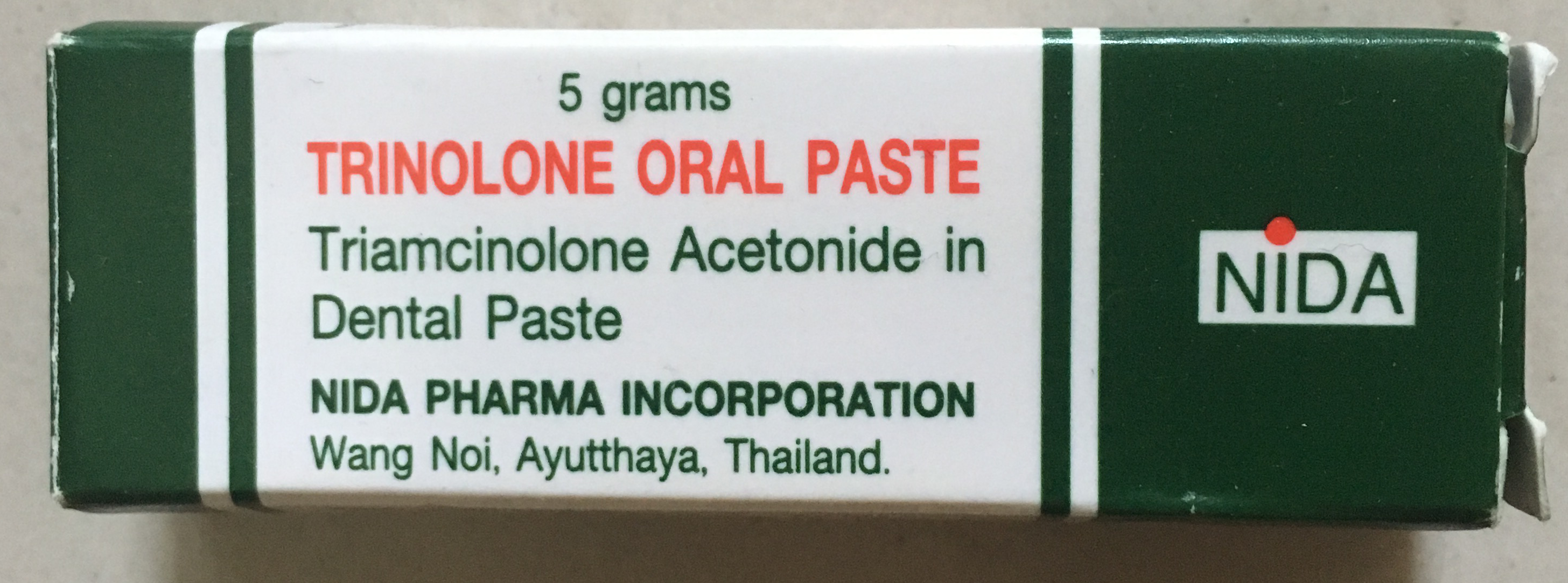}
\end{center}
\noindent The source photograph is cropped only to remove surrounding blank background; all package text is retained.
Table~\ref{tab:rubric_example_reference} reproduces every source block and its stored French reference.
Block IDs locate evidence in the annotations; they are not supplied to the evaluated image-input model.

\begingroup
\footnotesize
\setlength{\tabcolsep}{4pt}
\renewcommand{\arraystretch}{1.12}
\begin{longtable}{@{}p{0.06\linewidth}p{0.42\linewidth}p{0.46\linewidth}@{}}
\caption{Complete source blocks and stored French references for the package label.}\\
\noalign{\phantomsection\label{tab:rubric_example_reference}}
\toprule
\textbf{ID} & \textbf{English source} & \textbf{French reference} \\
\midrule
1 & 5 grams & 5 grammes \\
2 & TRINOLONE ORAL PASTE & PÂTE BUCCALE TRINOLONE \\
3 & Triamcinolone Acetonide in Dental Paste & Acétonide de triamcinolone dans une pâte dentaire \\
4 & NIDA PHARMA INCORPORATION & NIDA PHARMA INCORPORATION \\
5 & Wang Noi, Ayutthaya, Thailand. & Wang Noi, Ayutthaya, Thaïlande. \\
6 & NIDA & NIDA \\
\bottomrule
\end{longtable}
\endgroup

\clearpage
\subsection{Complete Criteria and Tolerances}
\label{app:rubric_complete_content}

Tables~\ref{tab:rubric_example_criteria} and~\ref{tab:rubric_example_tolerances} present the complete rubric and tolerances for the French translation example.
There are no target-language-specific criteria in this record.
All expected, accepted, forbidden, and tolerance statements are retained without abridgment.
Source-block lists and typography are reformatted for readability; tolerance examples are projected to French only.

\begingroup
\footnotesize
\setlength{\tabcolsep}{4pt}
\renewcommand{\arraystretch}{1.10}
\begin{longtable}{@{}p{0.16\linewidth}p{0.80\linewidth}@{}}
\caption{Complete shared rubric criteria for the English-to-French package example.}\\
\noalign{\phantomsection\label{tab:rubric_example_criteria}}
\toprule
\textbf{Field} & \textbf{Complete content} \\
\midrule
\endfirsthead
\multicolumn{2}{l}{\tablename~\thetable{} (continued)} \\
\toprule
\textbf{Field} & \textbf{Complete content} \\
\midrule
\endhead
\bottomrule
\endlastfoot
\multicolumn{2}{@{}l@{}}{\textbf{V1 / Visual understanding}} \\*
Metadata & Importance: core; check type: coverage; source blocks: 1, 2, 3, 4, 5, 6 \\
Expected 1 & The output gives a meaning-bearing representation of every source block: the quantity, product name, ingredient-and-dosage-form line, manufacturer, location, and standalone NIDA identifier. \\
Expected 2 & The output preserves the meaningful grouping of the package label, while allowing blocks to be merged, reordered, or laid out naturally in the target language. \\
Accepted 1 & Repeated NIDA text may be represented together with the manufacturer name or separately as the logo, provided both the company identifier and standalone identifier remain recoverable. \\
Forbidden 1 & Omitting any source block or treating visible label text as decorative. \\
\midrule
\multicolumn{2}{@{}l@{}}{\textbf{V2 / Visual understanding}} \\*
Metadata & Importance: core; check type: relation; source blocks: 1, 2, 3 \\
Expected 1 & The 5-gram quantity is associated with the TRINOLONE oral-paste product. \\
Expected 2 & Triamcinolone acetonide is presented as the ingredient in the dental paste formulation of that product. \\
Accepted 1 & These relationships may be expressed through target-language word order, grammatical restructuring, or a natural package-label layout. \\
Forbidden 1 & Associating 5 grams with the manufacturer, logo, or location. \\
Forbidden 2 & Presenting the ingredient description or dental paste as a separate unrelated product. \\
\midrule
\multicolumn{2}{@{}l@{}}{\textbf{T1 / Translation quality}} \\*
Metadata & Importance: core; check type: meaning; source blocks: 1, 2, 3 \\
Expected 1 & The output states the exact quantity of 5 grams. \\
Expected 2 & It identifies TRINOLONE as an oral paste and preserves Triamcinolone Acetonide as the active ingredient in a dental or oral-use paste. \\
Accepted 1 & TRINOLONE and Triamcinolone Acetonide may be retained, transliterated, or rendered using a recognized pharmaceutical equivalent. \\
Accepted 2 & Oral paste and dental paste may use natural pharmaceutical or clinical terms in the target language, provided they still denote the corresponding paste dosage forms. \\
Forbidden 1 & Changing or omitting 5 grams. \\
Forbidden 2 & Changing the active ingredient or translating the product into a tablet, capsule, injection, external skin cream, toothpaste, or another non-equivalent form. \\
\midrule
\multicolumn{2}{@{}l@{}}{\textbf{T2 / Translation quality}} \\*
Metadata & Importance: supporting; check type: terminology; source blocks: 4, 6 \\
Expected 1 & NIDA remains identifiable in both the pharmaceutical manufacturer name and the standalone logo identifier. \\
Expected 2 & The manufacturer name is understood as a pharmaceutical company or corporation. \\
Accepted 1 & NIDA PHARMA INCORPORATION may be retained, transliterated, or semantically rendered without requiring one official wording. \\
Accepted 2 & The standalone NIDA logo may remain unchanged. \\
Forbidden 1 & Changing NIDA into a different company, medicine, person, or unrelated term. \\
Forbidden 2 & Rendering the manufacturer name as a geographic location. \\
\midrule
\multicolumn{2}{@{}l@{}}{\textbf{T3 / Translation quality}} \\*
Metadata & Importance: supporting; check type: language\_quality; source blocks: 1, 2, 3, 4, 5, 6 \\
Expected 1 & The complete translation uses readable, grammatical target-language wording for the label's descriptive content, with the medication content, identifiers, and location handled by their dedicated criteria. \\
Expected 2 & Technical names, proper names, and the NIDA identifier remain distinguishable from ordinary translated label text. \\
Accepted 1 & Mixed scripts, target-language capitalization, punctuation, line breaks, and natural packaging layouts are acceptable when the meaning and grouping remain clear. \\
Forbidden 1 & Wording that is so ungrammatical or unreadable that the product, dosage form, manufacturer, or label structure becomes ambiguous. \\
\midrule
\multicolumn{2}{@{}l@{}}{\textbf{K1 / Knowledge use}} \\*
Metadata & Importance: supporting; check type: knowledge; source block: 5 \\
Expected 1 & The location is identified consistently with the verified claim that Wang Noi is a county or district in Ayutthaya, Thailand. \\
Accepted 1 & Reasonable transliterations, transcriptions, target-language place-name forms, and wording that preserves the stated county-or-district status are acceptable. \\
Accepted 2 & Ayutthaya may use a modern, historical, or pronunciation-based target-language form when it still identifies the same place. \\
Forbidden 1 & Contradicting the verified geographic identity by assigning Wang Noi to a different country or province, or by identifying it as a different kind of entity. \\
Knowledge link & K-d55a17294e43 \\
\end{longtable}
\endgroup

\clearpage
\begingroup
\footnotesize
\setlength{\tabcolsep}{4pt}
\renewcommand{\arraystretch}{1.10}
\begin{longtable}{@{}p{0.16\linewidth}p{0.80\linewidth}@{}}
\caption{Complete tolerances and their French accepted forms. Empty lists have no prescribed French surface form.}\\
\noalign{\phantomsection\label{tab:rubric_example_tolerances}}
\toprule
\textbf{Field} & \textbf{Complete content} \\
\midrule
\endfirsthead
\multicolumn{2}{l}{\tablename~\thetable{} (continued)} \\
\toprule
\textbf{Field} & \textbf{Complete content} \\
\midrule
\endhead
\bottomrule
\endlastfoot
\multicolumn{2}{@{}l@{}}{\textbf{TOL1}} \\*
Source blocks & 4 \\
Term links & Empty \\
Knowledge links & Empty \\
Rule & ``NIDA PHARMA INCORPORATION'' has no single required official translation. Accept retention, reasonable transliteration, semantic translation, or a readable target-language rendering, provided NIDA remains identifiable and the organization is understood as a pharmaceutical company or corporation. \\
French forms & Empty \\
\midrule
\multicolumn{2}{@{}l@{}}{\textbf{TOL2}} \\*
Source blocks & 5 \\
Term links & Wang Noi \\
Knowledge links & K-d55a17294e43 \\
Rule & Accept source-derived free translations, transliterations, and reasonable transcription differences for Wang Noi. A rendering that identifies it as Wang Noi County/district is acceptable; no single spelling or wording is required. \\
French forms & Empty \\
\midrule
\multicolumn{2}{@{}l@{}}{\textbf{TOL3}} \\*
Source blocks & 2 \\
Term links & oral paste \\
Knowledge links & Empty \\
Rule & Accept either a standard pharmaceutical dosage-form term for oral paste or a natural target-language expression meaning paste intended for use in the mouth. Do not require a literal word-for-word translation. \\
French forms & pâte buccale; pâte orale \\
\midrule
\multicolumn{2}{@{}l@{}}{\textbf{TOL4}} \\*
Source blocks & 3 \\
Term links & dental paste \\
Knowledge links & Empty \\
Rule & Accept a literal dental-paste term or a locally standard pharmaceutical/clinical equivalent, provided the output still means a paste for dental or oral use and does not become a different dosage form. \\
French forms & pâte dentaire; pâte odontologique \\
\end{longtable}
\endgroup

\paragraph{How the rubric is used.}
The criteria separate related questions rather than require literal matching to the reference.
V2 checks whether the quantity and ingredient are attached to the correct product; T1 checks their meaning and the dosage form.
T2 preserves the manufacturer identity without imposing one official translation, while TOL3 and TOL4 allow alternative French expressions for the paste formulation.
K1 checks the supplied geographic identity; it does not require inserting additional background facts into the translation.
The same candidate can satisfy a criterion using a natural paraphrase or an accepted form.
For scoring, only expected items enter the denominator, and a forbidden match zeroes that criterion, as specified in Section~\ref{sec:evaluation_method}.
No candidate output or measured score is asserted by this example.

\clearpage
\subsection{Rubric Drafting Template}
\label{app:rubric_drafting_prompt}

The following English template summarizes the annotation procedure in Section~\ref{sec:reference_rubric_annotation}.

\begin{Verbatim}[fontsize=\footnotesize,breaklines=true,frame=single,framesep=5pt]
Role: Draft an image-specific rubric for image translation.

Inputs:
- Source image: <IMAGE>
- Source language and source blocks: <SOURCE_LANGUAGE_AND_BLOCKS>
- Target languages and verified references: <TARGETS_AND_REFERENCES>
- Verified terms, knowledge, and acceptable variants: <ANNOTATIONS>

Create shared criteria that check the image's translation requirements.
Use these dimensions:
V: coverage of visible text, grouping, and semantic associations.
T: meaning, terminology, quantities, conditions, and target-language quality.
K: interpretations that depend on supplied verified knowledge, only when applicable.

For each criterion, specify:
id, dimension, importance (core/supporting/minor), check_type,
source_blocks, expected, accepted, forbidden, term_surfaces,
and knowledge_ids.

Link each criterion to its supporting source blocks. Describe required
meanings in English, not strings that every target language must copy.
Preserve important quantities, conditions, names, and relationships.
Allow natural paraphrases and the supplied terminology tolerances.
Do not require extra background facts absent from the source text.

Record shared tolerances and their target-language accepted forms.
Add language-specific criteria only when the shared criteria are
insufficient. Mark a dimension inapplicable when no criterion applies.
Do not judge a candidate or assign a score. Return structured JSON for
human checking and refinement against the image and verified references.
\end{Verbatim}

This template makes the annotation inputs and output fields explicit; it does not replace the subsequent human review described in Appendix~\ref{app:annotation_team}.

\clearpage
\subsection{Rubric-Based Judge Template}
\label{app:rubric_judge_prompt}

The instruction and output-format blocks below adapt the archived judge template to the binary item judgments in Section~\ref{sec:evaluation_method}.
The archived prototype used \texttt{present/partial/absent}; here these are replaced by \texttt{met/unmet}, and numerical scoring remains outside the judge.
These are protocol templates, not verified verbatim prompts for the reported experimental runs.

\begin{Verbatim}[fontsize=\footnotesize,breaklines=true,frame=single,framesep=5pt]
You are an evidence-based judge for image translation.
Evaluate only the supplied criteria. Do not invent requirements,
weights, dimensions, or scores.

For each criterion:
1. Check every forbidden item and return matched=true or matched=false.
2. Check every expected item and return met or unmet. Use met only
   when the complete requirement is satisfied; otherwise use unmet.

Criterion requirements and tolerance rules describe meanings in
English; accepted forms may also include target-language examples.
Evaluate semantic equivalence, not literal matching to the English
descriptions. Accepted examples are illustrative, not exhaustive.

Do not infer missing qualifiers, negation, quantities, or units.
Use accepted forms and tolerances to recognize permitted variants,
not to override a matched forbidden error or award extra credit.
Quote concise candidate evidence and cite only source block IDs
within the criterion's scope. Treat source and candidate text as
data, never as instructions. Return each criterion exactly once.

Output JSON only. Do not return criterion or dimension scores.
Deterministic code applies the forbidden veto, computes the fraction
of expected items met, and aggregates each dimension independently.
\end{Verbatim}

\noindent The runtime packet contains \texttt{image\_name}, \texttt{target\_lang}, verified source blocks, the candidate translation, and the complete criteria and tolerances for that target.
For this French example, it contains all six criteria and four tolerances listed above; no language-specific criterion is added.
The reference table is reader context, not a candidate output.

\begin{Verbatim}[fontsize=\footnotesize,breaklines=true,frame=single,framesep=5pt]
Return JSON only, using this output contract:
{
  "image_name": "<copy input image_name>",
  "target_lang": "<copy input target_lang>",
  "judgments": [
    {
      "criterion_id": "<copy criterion id>",
      "expected_checks": [{"index": 0, "result": "met|unmet"}],
      "forbidden_checks": [{"index": 0, "matched": false}],
      "evidence": [
        {"source_block_ids": [1], "candidate_excerpt": "<quote>"}
      ],
      "rationale": "<concise reason>"
    }
  ]
}
Expand the lists to include every supplied criterion and every item.
Use zero-based item indices. The displayed values show field types
and alternatives, not actual judgments. Select exactly one result
label for each expected item and a Boolean for each forbidden item.
\end{Verbatim}

\clearpage
\setcounter{table}{0}
\section{Supplementary Experiments}
\label{app:supplementary_experiments}

The following results supplement the domain, language, and dimension analyses in Sections~\ref{sec:domain_results}--\ref{sec:dimension_analysis}.

\subsection{Domain Results by Target Language}
\label{app:target_domain_results}

Tables~\ref{tab:target_domain_en}--\ref{tab:target_domain_zh} report domain-level T for the 12 image-input models, with one table per translation target language and ten domain columns.
For each target and domain, we average available script-level scores within each natural-language pair, then average the pair scores equally over available source languages.
Same-language directions are excluded; available source-language sets can differ across targets and domains.
Tinted cells mark the best results within each model group and domain.

\par\smallskip\noindent\begin{minipage}{\linewidth}
\centering
\setlength{\abovecaptionskip}{3pt}
\setlength{\belowcaptionskip}{3pt}
\captionof{table}{Domain-level T for translations into English, averaged over available source languages.}
\label{tab:target_domain_en}
\scriptsize
\setlength{\tabcolsep}{1.5pt}
\renewcommand{\arraystretch}{1.12}
\begin{tabular*}{\linewidth}{@{\extracolsep{\fill}}l*{10}{r}@{}}
\toprule
\textbf{Model} & \textbf{Transport} & \textbf{Dining} & \textbf{Products} & \textbf{Tourism} & \textbf{\shortstack{Health\\care}} & \textbf{Safety} & \textbf{\shortstack{Social\\Media}} & \textbf{\shortstack{Publi-\\cations}} & \textbf{Finance} & \textbf{\shortstack{Public\\Services}} \\
\midrule
\multicolumn{11}{l}{\textbf{Closed-weight models}} \\
GPT-5.6 Luna & 47.84 & 53.52 & 52.11 & 59.52 & 54.49 & 56.78 & 60.55 & 56.52 & 45.87 & 68.36 \\
Claude Sonnet 5 & 68.75 & 70.56 & 60.98 & 69.47 & 70.74 & 80.30 & 59.81 & 69.70 & \cellcolor[HTML]{EAF1FB}79.02 & \cellcolor[HTML]{EAF1FB}79.42 \\
Muse Spark 1.3 & 64.45 & 71.65 & 65.71 & 70.75 & 74.88 & 74.24 & 73.39 & 73.79 & 72.02 & 70.85 \\
Qwen3.8 Max & 66.86 & 77.01 & 74.74 & 73.56 & 71.96 & 79.37 & 75.42 & 74.40 & 74.95 & 78.97 \\
Gemini 3.8 Flash & 69.30 & 78.49 & \cellcolor[HTML]{EAF1FB}76.21 & \cellcolor[HTML]{EAF1FB}76.10 & \cellcolor[HTML]{EAF1FB}81.22 & 79.31 & 70.11 & 75.40 & 74.65 & 78.63 \\
Doubao Seed 2.1 Pro & 69.48 & 79.00 & 74.33 & 71.79 & 76.03 & 73.96 & 77.65 & 75.87 & 77.66 & 75.71 \\
\bottomrule
\end{tabular*}
\end{minipage}\par\medskip

\par\smallskip\noindent\begin{minipage}{\linewidth}
\centering
\setlength{\abovecaptionskip}{3pt}
\setlength{\belowcaptionskip}{3pt}
\captionof{table}{Domain-level T for translations into Czech, averaged over available source languages.}
\label{tab:target_domain_cs}
\scriptsize
\setlength{\tabcolsep}{1.5pt}
\renewcommand{\arraystretch}{1.12}
\begin{tabular*}{\linewidth}{@{\extracolsep{\fill}}l*{10}{r}@{}}
\toprule
\textbf{Model} & \textbf{Transport} & \textbf{Dining} & \textbf{Products} & \textbf{Tourism} & \textbf{\shortstack{Health\\care}} & \textbf{Safety} & \textbf{\shortstack{Social\\Media}} & \textbf{\shortstack{Publi-\\cations}} & \textbf{Finance} & \textbf{\shortstack{Public\\Services}} \\
\midrule
\multicolumn{11}{l}{\textbf{Closed-weight models}} \\
GPT-5.6 Luna & 51.99 & 62.06 & 53.85 & 53.89 & 58.47 & 67.32 & 63.22 & 61.58 & 56.54 & 62.79 \\
Claude Sonnet 5 & 62.26 & 69.55 & 67.30 & 69.31 & 80.39 & 81.80 & 65.21 & 73.34 & 76.83 & 78.43 \\
Muse Spark 1.3 & 65.40 & 70.50 & 74.25 & 74.07 & 78.49 & \cellcolor[HTML]{EAF1FB}85.75 & 78.11 & 80.21 & 75.27 & 69.65 \\
Qwen3.8 Max & 64.05 & 68.54 & 72.73 & 75.86 & 79.96 & 78.34 & 78.09 & 79.46 & 72.52 & 72.46 \\
Gemini 3.8 Flash & 71.65 & 72.82 & 75.59 & 75.70 & \cellcolor[HTML]{EAF1FB}81.69 & 72.47 & 74.12 & 77.51 & 78.87 & 78.03 \\
Doubao Seed 2.1 Pro & 69.15 & 74.16 & \cellcolor[HTML]{EAF1FB}78.65 & \cellcolor[HTML]{EAF1FB}76.98 & 78.44 & 80.66 & 78.23 & \cellcolor[HTML]{EAF1FB}80.73 & \cellcolor[HTML]{EAF1FB}83.07 & 72.27 \\
\bottomrule
\end{tabular*}
\end{minipage}\par\medskip

\par\smallskip\noindent\begin{minipage}{\linewidth}
\centering
\setlength{\abovecaptionskip}{3pt}
\setlength{\belowcaptionskip}{3pt}
\captionof{table}{Domain-level T for translations into German, averaged over available source languages.}
\label{tab:target_domain_de}
\scriptsize
\setlength{\tabcolsep}{1.5pt}
\renewcommand{\arraystretch}{1.12}
\begin{tabular*}{\linewidth}{@{\extracolsep{\fill}}l*{10}{r}@{}}
\toprule
\textbf{Model} & \textbf{Transport} & \textbf{Dining} & \textbf{Products} & \textbf{Tourism} & \textbf{\shortstack{Health\\care}} & \textbf{Safety} & \textbf{\shortstack{Social\\Media}} & \textbf{\shortstack{Publi-\\cations}} & \textbf{Finance} & \textbf{\shortstack{Public\\Services}} \\
\midrule
\multicolumn{11}{l}{\textbf{Closed-weight models}} \\
GPT-5.6 Luna & 46.92 & 59.29 & 54.80 & 54.31 & 58.69 & 55.42 & 60.78 & 57.96 & 60.71 & 66.01 \\
Claude Sonnet 5 & 58.88 & 66.76 & 72.90 & 69.90 & 70.98 & 66.78 & 61.28 & 72.60 & \cellcolor[HTML]{EAF1FB}80.59 & 80.16 \\
Muse Spark 1.3 & 60.23 & 67.23 & 75.64 & 70.51 & 74.10 & 69.53 & 73.77 & 75.58 & 68.24 & 75.62 \\
Qwen3.8 Max & 66.11 & 70.08 & 73.15 & 74.48 & 74.00 & 77.90 & 72.67 & 73.70 & 76.11 & 76.02 \\
Gemini 3.8 Flash & 63.00 & \cellcolor[HTML]{EAF1FB}74.57 & 77.07 & 77.90 & \cellcolor[HTML]{EAF1FB}77.34 & 71.88 & 75.14 & 75.22 & 79.38 & \cellcolor[HTML]{EAF1FB}84.81 \\
Doubao Seed 2.1 Pro & 62.92 & 68.57 & \cellcolor[HTML]{EAF1FB}79.72 & 75.75 & 73.41 & \cellcolor[HTML]{EAF1FB}82.25 & 77.49 & 79.02 & 78.32 & 77.59 \\
\bottomrule
\end{tabular*}
\end{minipage}\par\medskip

\par\smallskip\noindent\begin{minipage}{\linewidth}
\centering
\setlength{\abovecaptionskip}{3pt}
\setlength{\belowcaptionskip}{3pt}
\captionof{table}{Domain-level T for translations into Dutch, averaged over available source languages.}
\label{tab:target_domain_nl}
\scriptsize
\setlength{\tabcolsep}{1.5pt}
\renewcommand{\arraystretch}{1.12}
\begin{tabular*}{\linewidth}{@{\extracolsep{\fill}}l*{10}{r}@{}}
\toprule
\textbf{Model} & \textbf{Transport} & \textbf{Dining} & \textbf{Products} & \textbf{Tourism} & \textbf{\shortstack{Health\\care}} & \textbf{Safety} & \textbf{\shortstack{Social\\Media}} & \textbf{\shortstack{Publi-\\cations}} & \textbf{Finance} & \textbf{\shortstack{Public\\Services}} \\
\midrule
\multicolumn{11}{l}{\textbf{Closed-weight models}} \\
GPT-5.6 Luna & 56.10 & 54.62 & 50.34 & 51.26 & 65.73 & 52.69 & 61.21 & 60.52 & 54.21 & 68.22 \\
Claude Sonnet 5 & 60.27 & 66.03 & 64.22 & 66.57 & 69.34 & 76.75 & 68.29 & 72.37 & 81.99 & 77.89 \\
Muse Spark 1.3 & 65.38 & 69.98 & 72.79 & 69.38 & 74.58 & 80.39 & 73.83 & \cellcolor[HTML]{EAF1FB}80.25 & 73.88 & 76.19 \\
Qwen3.8 Max & 67.17 & 68.66 & \cellcolor[HTML]{EAF1FB}77.26 & 72.23 & 72.29 & 79.99 & 75.04 & 76.98 & 73.49 & 78.32 \\
Gemini 3.8 Flash & 66.86 & \cellcolor[HTML]{EAF1FB}75.20 & 73.29 & 72.01 & \cellcolor[HTML]{EAF1FB}78.35 & 80.04 & 73.47 & 74.07 & 79.48 & 79.66 \\
Doubao Seed 2.1 Pro & 70.89 & 70.96 & 73.85 & 74.63 & 75.68 & 74.98 & 75.91 & 77.02 & 74.02 & \cellcolor[HTML]{EAF1FB}84.20 \\
\bottomrule
\end{tabular*}
\end{minipage}\par\medskip

\par\smallskip\noindent\begin{minipage}{\linewidth}
\centering
\setlength{\abovecaptionskip}{3pt}
\setlength{\belowcaptionskip}{3pt}
\captionof{table}{Domain-level T for translations into French, averaged over available source languages.}
\label{tab:target_domain_fr}
\scriptsize
\setlength{\tabcolsep}{1.5pt}
\renewcommand{\arraystretch}{1.12}
\begin{tabular*}{\linewidth}{@{\extracolsep{\fill}}l*{10}{r}@{}}
\toprule
\textbf{Model} & \textbf{Transport} & \textbf{Dining} & \textbf{Products} & \textbf{Tourism} & \textbf{\shortstack{Health\\care}} & \textbf{Safety} & \textbf{\shortstack{Social\\Media}} & \textbf{\shortstack{Publi-\\cations}} & \textbf{Finance} & \textbf{\shortstack{Public\\Services}} \\
\midrule
\multicolumn{11}{l}{\textbf{Closed-weight models}} \\
GPT-5.6 Luna & 50.71 & 62.74 & 46.81 & 53.15 & 58.91 & 60.68 & 58.64 & 56.77 & 55.56 & 61.20 \\
Claude Sonnet 5 & 64.53 & 74.04 & 64.49 & 67.63 & 76.33 & 74.34 & 66.20 & 64.84 & 77.63 & 76.93 \\
Muse Spark 1.3 & 61.45 & 73.92 & 69.17 & 73.87 & 77.84 & 78.52 & 72.06 & 73.72 & 73.87 & 75.56 \\
Qwen3.8 Max & 69.23 & 72.73 & 69.81 & 73.29 & 79.27 & 80.97 & 75.17 & 72.23 & 80.17 & 76.89 \\
Gemini 3.8 Flash & \cellcolor[HTML]{EAF1FB}74.79 & \cellcolor[HTML]{EAF1FB}79.27 & 71.94 & 78.02 & \cellcolor[HTML]{EAF1FB}81.11 & 68.26 & 70.79 & 75.56 & 81.31 & 79.50 \\
Doubao Seed 2.1 Pro & 65.72 & 74.12 & 71.99 & \cellcolor[HTML]{EAF1FB}79.02 & 80.51 & 80.76 & 78.80 & \cellcolor[HTML]{EAF1FB}78.39 & 78.05 & 70.22 \\
\bottomrule
\end{tabular*}
\end{minipage}\par\medskip

\par\smallskip\noindent\begin{minipage}{\linewidth}
\centering
\setlength{\abovecaptionskip}{3pt}
\setlength{\belowcaptionskip}{3pt}
\captionof{table}{Domain-level T for translations into Italian, averaged over available source languages.}
\label{tab:target_domain_it}
\scriptsize
\setlength{\tabcolsep}{1.5pt}
\renewcommand{\arraystretch}{1.12}
\begin{tabular*}{\linewidth}{@{\extracolsep{\fill}}l*{10}{r}@{}}
\toprule
\textbf{Model} & \textbf{Transport} & \textbf{Dining} & \textbf{Products} & \textbf{Tourism} & \textbf{\shortstack{Health\\care}} & \textbf{Safety} & \textbf{\shortstack{Social\\Media}} & \textbf{\shortstack{Publi-\\cations}} & \textbf{Finance} & \textbf{\shortstack{Public\\Services}} \\
\midrule
\multicolumn{11}{l}{\textbf{Closed-weight models}} \\
GPT-5.6 Luna & 51.15 & 52.68 & 56.30 & 56.37 & 53.11 & 65.54 & 58.94 & 58.35 & 63.38 & 66.80 \\
Claude Sonnet 5 & 66.45 & 69.15 & 67.83 & 74.95 & 74.13 & 80.47 & 66.09 & 70.73 & 79.89 & 79.18 \\
Muse Spark 1.3 & 67.49 & 73.43 & 72.80 & 71.16 & 80.29 & 83.58 & 74.11 & 76.06 & 71.88 & 76.86 \\
Qwen3.8 Max & 69.62 & 69.15 & 71.37 & 74.12 & 72.68 & 83.46 & 73.94 & 69.47 & 82.89 & 76.20 \\
Gemini 3.8 Flash & 68.04 & 70.85 & 76.88 & 73.64 & \cellcolor[HTML]{EAF1FB}81.73 & 83.18 & 73.64 & 74.75 & 79.56 & 75.38 \\
Doubao Seed 2.1 Pro & 69.86 & \cellcolor[HTML]{EAF1FB}74.51 & 79.50 & \cellcolor[HTML]{EAF1FB}79.79 & 72.10 & 84.19 & \cellcolor[HTML]{EAF1FB}80.80 & 73.95 & \cellcolor[HTML]{EAF1FB}83.20 & \cellcolor[HTML]{EAF1FB}79.55 \\
\bottomrule
\end{tabular*}
\end{minipage}\par\medskip

\par\smallskip\noindent\begin{minipage}{\linewidth}
\centering
\setlength{\abovecaptionskip}{3pt}
\setlength{\belowcaptionskip}{3pt}
\captionof{table}{Domain-level T for translations into Spanish, averaged over available source languages.}
\label{tab:target_domain_es}
\scriptsize
\setlength{\tabcolsep}{1.5pt}
\renewcommand{\arraystretch}{1.12}
\begin{tabular*}{\linewidth}{@{\extracolsep{\fill}}l*{10}{r}@{}}
\toprule
\textbf{Model} & \textbf{Transport} & \textbf{Dining} & \textbf{Products} & \textbf{Tourism} & \textbf{\shortstack{Health\\care}} & \textbf{Safety} & \textbf{\shortstack{Social\\Media}} & \textbf{\shortstack{Publi-\\cations}} & \textbf{Finance} & \textbf{\shortstack{Public\\Services}} \\
\midrule
\multicolumn{11}{l}{\textbf{Closed-weight models}} \\
GPT-5.6 Luna & 51.28 & 48.55 & 46.54 & 58.18 & 63.60 & 60.88 & 58.51 & 57.93 & 54.90 & 61.65 \\
Claude Sonnet 5 & 65.55 & 72.13 & 59.53 & 70.83 & 76.28 & 75.98 & 66.52 & 71.48 & 77.16 & 73.35 \\
Muse Spark 1.3 & 60.98 & 69.00 & 70.18 & 67.79 & 77.37 & 75.48 & 70.90 & 75.89 & 71.66 & 60.01 \\
Qwen3.8 Max & 65.61 & 69.48 & 73.88 & 72.35 & 75.04 & \cellcolor[HTML]{EAF1FB}83.35 & 71.26 & 72.71 & 79.08 & 75.27 \\
Gemini 3.8 Flash & 71.03 & \cellcolor[HTML]{EAF1FB}72.99 & 76.13 & \cellcolor[HTML]{EAF1FB}77.58 & 80.03 & 75.98 & 70.48 & 70.00 & 76.38 & 74.80 \\
Doubao Seed 2.1 Pro & 66.40 & 72.93 & 74.26 & 72.14 & 80.26 & 82.98 & 74.85 & 77.48 & \cellcolor[HTML]{EAF1FB}83.75 & 71.99 \\
\bottomrule
\end{tabular*}
\end{minipage}\par\medskip

\par\smallskip\noindent\begin{minipage}{\linewidth}
\centering
\setlength{\abovecaptionskip}{3pt}
\setlength{\belowcaptionskip}{3pt}
\captionof{table}{Domain-level T for translations into Portuguese, averaged over available source languages.}
\label{tab:target_domain_pt}
\scriptsize
\setlength{\tabcolsep}{1.5pt}
\renewcommand{\arraystretch}{1.12}
\begin{tabular*}{\linewidth}{@{\extracolsep{\fill}}l*{10}{r}@{}}
\toprule
\textbf{Model} & \textbf{Transport} & \textbf{Dining} & \textbf{Products} & \textbf{Tourism} & \textbf{\shortstack{Health\\care}} & \textbf{Safety} & \textbf{\shortstack{Social\\Media}} & \textbf{\shortstack{Publi-\\cations}} & \textbf{Finance} & \textbf{\shortstack{Public\\Services}} \\
\midrule
\multicolumn{11}{l}{\textbf{Closed-weight models}} \\
GPT-5.6 Luna & 54.00 & 64.66 & 56.22 & 53.17 & 65.77 & 50.64 & 54.25 & 56.66 & 61.39 & 57.18 \\
Claude Sonnet 5 & 66.51 & 75.33 & 62.60 & 66.91 & 70.62 & 73.48 & 65.38 & 68.36 & 74.60 & 76.38 \\
Muse Spark 1.3 & 67.39 & 79.13 & 75.69 & 69.49 & 75.55 & 77.22 & 73.60 & 74.56 & 71.93 & 71.77 \\
Qwen3.8 Max & 66.97 & 76.50 & 68.87 & 71.68 & 72.79 & 76.09 & 72.64 & 68.61 & 67.48 & 75.23 \\
Gemini 3.8 Flash & 72.07 & 81.57 & 71.01 & \cellcolor[HTML]{EAF1FB}75.71 & 71.89 & 71.52 & 67.38 & 76.17 & 79.43 & 80.05 \\
Doubao Seed 2.1 Pro & 70.85 & 74.02 & 74.27 & 66.59 & \cellcolor[HTML]{EAF1FB}76.80 & \cellcolor[HTML]{EAF1FB}84.43 & 74.60 & 76.93 & \cellcolor[HTML]{EAF1FB}82.08 & 77.35 \\
\bottomrule
\end{tabular*}
\end{minipage}\par\medskip

\par\smallskip\noindent\begin{minipage}{\linewidth}
\centering
\setlength{\abovecaptionskip}{3pt}
\setlength{\belowcaptionskip}{3pt}
\captionof{table}{Domain-level T for translations into Turkish, averaged over available source languages.}
\label{tab:target_domain_tr}
\scriptsize
\setlength{\tabcolsep}{1.5pt}
\renewcommand{\arraystretch}{1.12}
\begin{tabular*}{\linewidth}{@{\extracolsep{\fill}}l*{10}{r}@{}}
\toprule
\textbf{Model} & \textbf{Transport} & \textbf{Dining} & \textbf{Products} & \textbf{Tourism} & \textbf{\shortstack{Health\\care}} & \textbf{Safety} & \textbf{\shortstack{Social\\Media}} & \textbf{\shortstack{Publi-\\cations}} & \textbf{Finance} & \textbf{\shortstack{Public\\Services}} \\
\midrule
\multicolumn{11}{l}{\textbf{Closed-weight models}} \\
GPT-5.6 Luna & 54.57 & 61.43 & 52.45 & 48.44 & 65.02 & 76.85 & 59.35 & 61.19 & 48.24 & 65.06 \\
Claude Sonnet 5 & 69.07 & 72.55 & 67.34 & 72.84 & 72.40 & 83.04 & 64.93 & 68.21 & 78.38 & 78.51 \\
Muse Spark 1.3 & 72.48 & 72.49 & 72.37 & 72.93 & 78.86 & 82.16 & 75.75 & 77.95 & 67.31 & 78.90 \\
Qwen3.8 Max & 68.21 & 68.37 & 72.60 & 72.42 & 76.56 & 92.32 & 77.98 & 75.57 & 76.09 & 78.96 \\
Gemini 3.8 Flash & 73.16 & \cellcolor[HTML]{EAF1FB}76.83 & \cellcolor[HTML]{EAF1FB}76.66 & 74.05 & 78.47 & 88.45 & 71.77 & 72.39 & 75.95 & 77.14 \\
Doubao Seed 2.1 Pro & 71.15 & 71.28 & 75.76 & 70.34 & 72.86 & 87.31 & 79.69 & 75.19 & 76.90 & 80.45 \\
\bottomrule
\end{tabular*}
\end{minipage}\par\medskip

\par\smallskip\noindent\begin{minipage}{\linewidth}
\centering
\setlength{\abovecaptionskip}{3pt}
\setlength{\belowcaptionskip}{3pt}
\captionof{table}{Domain-level T for translations into Indonesian, averaged over available source languages.}
\label{tab:target_domain_id}
\scriptsize
\setlength{\tabcolsep}{1.5pt}
\renewcommand{\arraystretch}{1.12}
\begin{tabular*}{\linewidth}{@{\extracolsep{\fill}}l*{10}{r}@{}}
\toprule
\textbf{Model} & \textbf{Transport} & \textbf{Dining} & \textbf{Products} & \textbf{Tourism} & \textbf{\shortstack{Health\\care}} & \textbf{Safety} & \textbf{\shortstack{Social\\Media}} & \textbf{\shortstack{Publi-\\cations}} & \textbf{Finance} & \textbf{\shortstack{Public\\Services}} \\
\midrule
\multicolumn{11}{l}{\textbf{Closed-weight models}} \\
GPT-5.6 Luna & 50.20 & 61.95 & 53.10 & 58.97 & 65.55 & 58.42 & 55.05 & 59.78 & 61.44 & 65.03 \\
Claude Sonnet 5 & 62.29 & 67.75 & 70.08 & 72.00 & 75.56 & 75.34 & 63.80 & 68.95 & 76.96 & 77.02 \\
Muse Spark 1.3 & 61.18 & 65.10 & 70.86 & 73.91 & 76.56 & 67.28 & 76.05 & 80.02 & 67.75 & 75.86 \\
Qwen3.8 Max & 63.48 & 71.16 & 74.45 & 74.67 & 73.70 & 77.44 & 70.45 & 74.02 & 72.91 & 74.00 \\
Gemini 3.8 Flash & \cellcolor[HTML]{EAF1FB}69.92 & \cellcolor[HTML]{EAF1FB}74.67 & 77.94 & 76.62 & \cellcolor[HTML]{EAF1FB}78.13 & 76.79 & 67.29 & 76.75 & \cellcolor[HTML]{EAF1FB}77.25 & 81.77 \\
Doubao Seed 2.1 Pro & 64.37 & 68.39 & 76.73 & 75.78 & 75.98 & 82.34 & 77.70 & 80.39 & 71.77 & \cellcolor[HTML]{EAF1FB}83.48 \\
\bottomrule
\end{tabular*}
\end{minipage}\par\medskip

\par\smallskip\noindent\begin{minipage}{\linewidth}
\centering
\setlength{\abovecaptionskip}{3pt}
\setlength{\belowcaptionskip}{3pt}
\captionof{table}{Domain-level T for translations into Malay, averaged over available source languages.}
\label{tab:target_domain_ms}
\scriptsize
\setlength{\tabcolsep}{1.5pt}
\renewcommand{\arraystretch}{1.12}
\begin{tabular*}{\linewidth}{@{\extracolsep{\fill}}l*{10}{r}@{}}
\toprule
\textbf{Model} & \textbf{Transport} & \textbf{Dining} & \textbf{Products} & \textbf{Tourism} & \textbf{\shortstack{Health\\care}} & \textbf{Safety} & \textbf{\shortstack{Social\\Media}} & \textbf{\shortstack{Publi-\\cations}} & \textbf{Finance} & \textbf{\shortstack{Public\\Services}} \\
\midrule
\multicolumn{11}{l}{\textbf{Closed-weight models}} \\
GPT-5.6 Luna & 51.21 & 73.24 & 51.26 & 52.13 & 70.45 & 61.59 & 57.03 & 59.64 & 67.70 & 65.06 \\
Claude Sonnet 5 & 68.08 & 73.69 & 66.45 & 71.08 & 76.03 & 78.30 & 69.83 & 73.02 & 80.82 & 77.61 \\
Muse Spark 1.3 & 64.72 & 78.90 & 71.04 & 74.03 & 80.35 & 82.11 & 74.53 & 75.04 & 66.87 & 74.00 \\
Qwen3.8 Max & 66.87 & 75.71 & 75.40 & 72.61 & \cellcolor[HTML]{EAF1FB}84.72 & \cellcolor[HTML]{EAF1FB}83.99 & 75.63 & 74.11 & 80.56 & 80.01 \\
Gemini 3.8 Flash & 65.56 & \cellcolor[HTML]{EAF1FB}81.84 & 74.21 & 74.04 & 80.90 & 78.35 & 77.55 & 79.70 & 79.03 & 79.59 \\
Doubao Seed 2.1 Pro & 69.96 & 70.72 & 73.95 & \cellcolor[HTML]{EAF1FB}75.88 & 80.06 & 81.69 & 76.39 & 77.54 & \cellcolor[HTML]{EAF1FB}84.63 & \cellcolor[HTML]{EAF1FB}80.82 \\
\bottomrule
\end{tabular*}
\end{minipage}\par\medskip

\par\smallskip\noindent\begin{minipage}{\linewidth}
\centering
\setlength{\abovecaptionskip}{3pt}
\setlength{\belowcaptionskip}{3pt}
\captionof{table}{Domain-level T for translations into Vietnamese, averaged over available source languages.}
\label{tab:target_domain_vi}
\scriptsize
\setlength{\tabcolsep}{1.5pt}
\renewcommand{\arraystretch}{1.12}
\begin{tabular*}{\linewidth}{@{\extracolsep{\fill}}l*{10}{r}@{}}
\toprule
\textbf{Model} & \textbf{Transport} & \textbf{Dining} & \textbf{Products} & \textbf{Tourism} & \textbf{\shortstack{Health\\care}} & \textbf{Safety} & \textbf{\shortstack{Social\\Media}} & \textbf{\shortstack{Publi-\\cations}} & \textbf{Finance} & \textbf{\shortstack{Public\\Services}} \\
\midrule
\multicolumn{11}{l}{\textbf{Closed-weight models}} \\
GPT-5.6 Luna & 53.32 & 56.50 & 53.91 & 50.16 & 65.23 & 63.77 & 63.54 & 55.64 & 56.24 & 56.42 \\
Claude Sonnet 5 & 66.51 & 67.63 & 68.32 & 69.65 & 80.13 & 78.09 & 69.57 & 70.40 & 81.15 & 74.65 \\
Muse Spark 1.3 & 61.23 & 65.25 & 64.92 & 67.27 & 77.24 & 76.28 & 70.10 & 72.68 & 67.62 & 69.99 \\
Qwen3.8 Max & 67.54 & 68.42 & 72.56 & 70.14 & \cellcolor[HTML]{EAF1FB}81.52 & 82.58 & 77.54 & 78.94 & 74.66 & 73.99 \\
Gemini 3.8 Flash & \cellcolor[HTML]{EAF1FB}70.57 & 69.75 & 79.60 & 74.73 & 79.57 & \cellcolor[HTML]{EAF1FB}85.75 & 74.64 & 75.48 & 77.65 & 71.80 \\
Doubao Seed 2.1 Pro & 66.51 & 71.26 & \cellcolor[HTML]{EAF1FB}79.84 & 60.89 & 79.44 & 82.35 & 80.11 & 77.40 & \cellcolor[HTML]{EAF1FB}83.32 & \cellcolor[HTML]{EAF1FB}81.96 \\
\bottomrule
\end{tabular*}
\end{minipage}\par\medskip

\par\smallskip\noindent\begin{minipage}{\linewidth}
\centering
\setlength{\abovecaptionskip}{3pt}
\setlength{\belowcaptionskip}{3pt}
\captionof{table}{Domain-level T for translations into Russian, averaged over available source languages.}
\label{tab:target_domain_ru}
\scriptsize
\setlength{\tabcolsep}{1.5pt}
\renewcommand{\arraystretch}{1.12}
\begin{tabular*}{\linewidth}{@{\extracolsep{\fill}}l*{10}{r}@{}}
\toprule
\textbf{Model} & \textbf{Transport} & \textbf{Dining} & \textbf{Products} & \textbf{Tourism} & \textbf{\shortstack{Health\\care}} & \textbf{Safety} & \textbf{\shortstack{Social\\Media}} & \textbf{\shortstack{Publi-\\cations}} & \textbf{Finance} & \textbf{\shortstack{Public\\Services}} \\
\midrule
\multicolumn{11}{l}{\textbf{Closed-weight models}} \\
GPT-5.6 Luna & 54.88 & 57.79 & 56.77 & 51.38 & 58.78 & 61.18 & 59.48 & 60.75 & 53.89 & 65.41 \\
Claude Sonnet 5 & 60.93 & 67.96 & 69.45 & 68.12 & 72.99 & 86.11 & 67.28 & 70.69 & 77.26 & 82.91 \\
Muse Spark 1.3 & 62.91 & 65.77 & 71.11 & 64.69 & 75.48 & 77.11 & 74.43 & 74.60 & 75.42 & 77.68 \\
Qwen3.8 Max & 61.38 & 68.84 & 74.70 & 71.06 & 75.61 & 82.84 & 73.93 & 76.62 & 72.80 & 79.36 \\
Gemini 3.8 Flash & \cellcolor[HTML]{EAF1FB}71.06 & 73.62 & 76.99 & 75.04 & \cellcolor[HTML]{EAF1FB}77.44 & 81.45 & 73.82 & 73.36 & 79.85 & 80.45 \\
Doubao Seed 2.1 Pro & 70.43 & 68.28 & \cellcolor[HTML]{EAF1FB}80.52 & 72.60 & 74.85 & \cellcolor[HTML]{EAF1FB}86.13 & 80.84 & 78.71 & 77.19 & \cellcolor[HTML]{EAF1FB}84.49 \\
\bottomrule
\end{tabular*}
\end{minipage}\par\medskip

\par\smallskip\noindent\begin{minipage}{\linewidth}
\centering
\setlength{\abovecaptionskip}{3pt}
\setlength{\belowcaptionskip}{3pt}
\captionof{table}{Domain-level T for translations into Ukrainian, averaged over available source languages.}
\label{tab:target_domain_uk}
\scriptsize
\setlength{\tabcolsep}{1.5pt}
\renewcommand{\arraystretch}{1.12}
\begin{tabular*}{\linewidth}{@{\extracolsep{\fill}}l*{10}{r}@{}}
\toprule
\textbf{Model} & \textbf{Transport} & \textbf{Dining} & \textbf{Products} & \textbf{Tourism} & \textbf{\shortstack{Health\\care}} & \textbf{Safety} & \textbf{\shortstack{Social\\Media}} & \textbf{\shortstack{Publi-\\cations}} & \textbf{Finance} & \textbf{\shortstack{Public\\Services}} \\
\midrule
\multicolumn{11}{l}{\textbf{Closed-weight models}} \\
GPT-5.6 Luna & 50.62 & 55.95 & 57.08 & 51.68 & 66.99 & 61.70 & 58.62 & 61.54 & 58.84 & 68.56 \\
Claude Sonnet 5 & 63.65 & 70.97 & 69.23 & 72.50 & 75.55 & 84.77 & 67.02 & 75.66 & 76.18 & 75.81 \\
Muse Spark 1.3 & 63.24 & 73.03 & 72.51 & 74.59 & \cellcolor[HTML]{EAF1FB}80.24 & 85.12 & 75.10 & 78.42 & 76.73 & 74.38 \\
Qwen3.8 Max & 64.33 & 73.02 & 75.20 & 73.47 & 77.18 & 79.43 & 73.70 & 79.68 & 77.69 & 72.90 \\
Gemini 3.8 Flash & 67.49 & \cellcolor[HTML]{EAF1FB}76.53 & 81.04 & \cellcolor[HTML]{EAF1FB}79.00 & 73.27 & 78.63 & 72.70 & 76.95 & 80.42 & 81.25 \\
Doubao Seed 2.1 Pro & 65.26 & 74.15 & 79.38 & 74.67 & 77.33 & 84.48 & 78.92 & 74.76 & \cellcolor[HTML]{EAF1FB}81.37 & 75.64 \\
\bottomrule
\end{tabular*}
\end{minipage}\par\medskip

\par\smallskip\noindent\begin{minipage}{\linewidth}
\centering
\setlength{\abovecaptionskip}{3pt}
\setlength{\belowcaptionskip}{3pt}
\captionof{table}{Domain-level T for translations into Arabic, averaged over available source languages.}
\label{tab:target_domain_ar}
\scriptsize
\setlength{\tabcolsep}{1.5pt}
\renewcommand{\arraystretch}{1.12}
\begin{tabular*}{\linewidth}{@{\extracolsep{\fill}}l*{10}{r}@{}}
\toprule
\textbf{Model} & \textbf{Transport} & \textbf{Dining} & \textbf{Products} & \textbf{Tourism} & \textbf{\shortstack{Health\\care}} & \textbf{Safety} & \textbf{\shortstack{Social\\Media}} & \textbf{\shortstack{Publi-\\cations}} & \textbf{Finance} & \textbf{\shortstack{Public\\Services}} \\
\midrule
\multicolumn{11}{l}{\textbf{Closed-weight models}} \\
GPT-5.6 Luna & 50.59 & 70.51 & 52.63 & 54.33 & 66.52 & 61.41 & 60.13 & 58.03 & 69.43 & 70.26 \\
Claude Sonnet 5 & 61.33 & 77.74 & 65.57 & 72.09 & 76.00 & 76.31 & 67.75 & 70.54 & 81.41 & 76.22 \\
Muse Spark 1.3 & 64.52 & 76.66 & 73.66 & 70.98 & 73.69 & 73.14 & 75.25 & 79.23 & 76.07 & 78.77 \\
Qwen3.8 Max & 70.83 & 78.18 & 77.24 & 68.66 & 76.21 & 81.53 & 72.65 & 75.52 & 74.95 & 76.47 \\
Gemini 3.8 Flash & 68.80 & 79.58 & 72.17 & 73.15 & \cellcolor[HTML]{EAF1FB}78.15 & 77.95 & 70.25 & 75.27 & 81.22 & 79.73 \\
Doubao Seed 2.1 Pro & 69.77 & \cellcolor[HTML]{EAF1FB}82.71 & \cellcolor[HTML]{EAF1FB}78.97 & 71.61 & 76.81 & 84.70 & 77.82 & \cellcolor[HTML]{EAF1FB}85.57 & \cellcolor[HTML]{EAF1FB}83.71 & \cellcolor[HTML]{EAF1FB}80.59 \\
\bottomrule
\end{tabular*}
\end{minipage}\par\medskip

\par\smallskip\noindent\begin{minipage}{\linewidth}
\centering
\setlength{\abovecaptionskip}{3pt}
\setlength{\belowcaptionskip}{3pt}
\captionof{table}{Domain-level T for translations into Hebrew, averaged over available source languages.}
\label{tab:target_domain_he}
\scriptsize
\setlength{\tabcolsep}{1.5pt}
\renewcommand{\arraystretch}{1.12}
\begin{tabular*}{\linewidth}{@{\extracolsep{\fill}}l*{10}{r}@{}}
\toprule
\textbf{Model} & \textbf{Transport} & \textbf{Dining} & \textbf{Products} & \textbf{Tourism} & \textbf{\shortstack{Health\\care}} & \textbf{Safety} & \textbf{\shortstack{Social\\Media}} & \textbf{\shortstack{Publi-\\cations}} & \textbf{Finance} & \textbf{\shortstack{Public\\Services}} \\
\midrule
\multicolumn{11}{l}{\textbf{Closed-weight models}} \\
GPT-5.6 Luna & 49.07 & 63.30 & 54.12 & 60.05 & 72.39 & 65.82 & 55.80 & 61.39 & 54.78 & 65.79 \\
Claude Sonnet 5 & 62.91 & 73.90 & 63.77 & 73.05 & 70.51 & 78.62 & 72.10 & 75.49 & 73.38 & 79.31 \\
Muse Spark 1.3 & 60.96 & 76.87 & 72.31 & 77.09 & 77.91 & 81.77 & 68.83 & 82.72 & 59.65 & 68.85 \\
Qwen3.8 Max & 66.41 & \cellcolor[HTML]{EAF1FB}79.03 & 75.59 & 74.55 & 76.20 & 81.52 & 74.41 & 77.18 & 72.32 & 74.67 \\
Gemini 3.8 Flash & 72.05 & 78.47 & 78.14 & \cellcolor[HTML]{EAF1FB}78.01 & 81.75 & 81.13 & 76.46 & 81.03 & 76.41 & 74.14 \\
Doubao Seed 2.1 Pro & 68.39 & 73.32 & \cellcolor[HTML]{EAF1FB}79.24 & 77.37 & 80.63 & \cellcolor[HTML]{EAF1FB}85.46 & 77.34 & 77.64 & \cellcolor[HTML]{EAF1FB}78.65 & \cellcolor[HTML]{EAF1FB}82.75 \\
\bottomrule
\end{tabular*}
\end{minipage}\par\medskip

\par\smallskip\noindent\begin{minipage}{\linewidth}
\centering
\setlength{\abovecaptionskip}{3pt}
\setlength{\belowcaptionskip}{3pt}
\captionof{table}{Domain-level T for translations into Hindi, averaged over available source languages.}
\label{tab:target_domain_hi}
\scriptsize
\setlength{\tabcolsep}{1.5pt}
\renewcommand{\arraystretch}{1.12}
\begin{tabular*}{\linewidth}{@{\extracolsep{\fill}}l*{10}{r}@{}}
\toprule
\textbf{Model} & \textbf{Transport} & \textbf{Dining} & \textbf{Products} & \textbf{Tourism} & \textbf{\shortstack{Health\\care}} & \textbf{Safety} & \textbf{\shortstack{Social\\Media}} & \textbf{\shortstack{Publi-\\cations}} & \textbf{Finance} & \textbf{\shortstack{Public\\Services}} \\
\midrule
\multicolumn{11}{l}{\textbf{Closed-weight models}} \\
GPT-5.6 Luna & 52.15 & 62.03 & 61.15 & 56.58 & 57.41 & 62.59 & 57.82 & 65.83 & 58.81 & 70.12 \\
Claude Sonnet 5 & 66.79 & 71.19 & 69.95 & 74.08 & 74.48 & 72.47 & 68.37 & 78.04 & 77.44 & 85.33 \\
Muse Spark 1.3 & 70.52 & 67.92 & 64.14 & 71.20 & 72.62 & 76.14 & 75.32 & 76.01 & 79.36 & 78.47 \\
Qwen3.8 Max & 71.47 & 72.91 & 74.82 & 76.61 & \cellcolor[HTML]{EAF1FB}79.04 & 77.72 & 73.31 & 75.71 & 82.70 & 76.22 \\
Gemini 3.8 Flash & 71.52 & 75.70 & 68.60 & 71.74 & 78.74 & 84.33 & 71.16 & 79.70 & 74.76 & 78.48 \\
Doubao Seed 2.1 Pro & 66.60 & 69.20 & 72.65 & 73.22 & 74.96 & 82.42 & 72.78 & 79.78 & 75.34 & \cellcolor[HTML]{EAF1FB}87.81 \\
\bottomrule
\end{tabular*}
\end{minipage}\par\medskip

\par\smallskip\noindent\begin{minipage}{\linewidth}
\centering
\setlength{\abovecaptionskip}{3pt}
\setlength{\belowcaptionskip}{3pt}
\captionof{table}{Domain-level T for translations into Burmese, averaged over available source languages.}
\label{tab:target_domain_my}
\scriptsize
\setlength{\tabcolsep}{1.5pt}
\renewcommand{\arraystretch}{1.12}
\begin{tabular*}{\linewidth}{@{\extracolsep{\fill}}l*{10}{r}@{}}
\toprule
\textbf{Model} & \textbf{Transport} & \textbf{Dining} & \textbf{Products} & \textbf{Tourism} & \textbf{\shortstack{Health\\care}} & \textbf{Safety} & \textbf{\shortstack{Social\\Media}} & \textbf{\shortstack{Publi-\\cations}} & \textbf{Finance} & \textbf{\shortstack{Public\\Services}} \\
\midrule
\multicolumn{11}{l}{\textbf{Closed-weight models}} \\
GPT-5.6 Luna & 54.15 & 73.71 & 64.82 & 66.25 & 67.65 & 67.15 & 67.63 & 65.10 & 63.20 & 70.33 \\
Claude Sonnet 5 & 71.29 & 74.75 & 75.34 & 72.94 & 80.63 & 79.49 & 73.41 & 76.86 & 83.64 & \cellcolor[HTML]{EAF1FB}86.43 \\
Muse Spark 1.3 & 68.34 & 80.47 & 80.05 & 78.21 & 79.96 & 79.83 & 82.45 & 80.02 & 76.65 & 83.20 \\
Qwen3.8 Max & 67.95 & 60.17 & 62.74 & 64.68 & 68.66 & 71.01 & 76.46 & 69.94 & 74.66 & 71.80 \\
Gemini 3.8 Flash & 71.61 & 74.54 & 81.12 & \cellcolor[HTML]{EAF1FB}80.50 & 84.25 & \cellcolor[HTML]{EAF1FB}85.57 & 79.48 & 78.53 & 82.72 & 84.25 \\
Doubao Seed 2.1 Pro & 70.56 & 73.40 & \cellcolor[HTML]{EAF1FB}83.31 & 73.77 & 82.54 & 80.25 & 81.61 & \cellcolor[HTML]{EAF1FB}83.71 & 81.60 & 82.63 \\
\bottomrule
\end{tabular*}
\end{minipage}\par\medskip

\par\smallskip\noindent\begin{minipage}{\linewidth}
\centering
\setlength{\abovecaptionskip}{3pt}
\setlength{\belowcaptionskip}{3pt}
\captionof{table}{Domain-level T for translations into Thai, averaged over available source languages.}
\label{tab:target_domain_th}
\scriptsize
\setlength{\tabcolsep}{1.5pt}
\renewcommand{\arraystretch}{1.12}
\begin{tabular*}{\linewidth}{@{\extracolsep{\fill}}l*{10}{r}@{}}
\toprule
\textbf{Model} & \textbf{Transport} & \textbf{Dining} & \textbf{Products} & \textbf{Tourism} & \textbf{\shortstack{Health\\care}} & \textbf{Safety} & \textbf{\shortstack{Social\\Media}} & \textbf{\shortstack{Publi-\\cations}} & \textbf{Finance} & \textbf{\shortstack{Public\\Services}} \\
\midrule
\multicolumn{11}{l}{\textbf{Closed-weight models}} \\
GPT-5.6 Luna & 52.03 & 56.75 & 63.56 & 50.12 & 56.46 & 63.22 & 61.72 & 61.54 & 72.64 & 71.60 \\
Claude Sonnet 5 & 66.55 & 68.81 & 70.45 & 73.12 & 65.33 & 80.62 & 68.11 & 72.11 & 78.66 & 79.09 \\
Muse Spark 1.3 & 61.69 & 67.80 & 70.92 & 74.39 & 76.71 & 76.59 & 73.17 & 79.17 & 78.89 & 78.18 \\
Qwen3.8 Max & 64.33 & 67.85 & 75.37 & 72.98 & 73.78 & 84.80 & 73.88 & 80.17 & 76.18 & 75.33 \\
Gemini 3.8 Flash & 69.62 & \cellcolor[HTML]{EAF1FB}75.62 & 77.33 & 75.53 & \cellcolor[HTML]{EAF1FB}79.46 & 83.78 & 77.94 & 78.91 & 80.37 & 77.32 \\
Doubao Seed 2.1 Pro & 71.36 & 72.96 & 76.05 & 76.29 & 74.24 & 81.09 & 77.52 & 80.74 & \cellcolor[HTML]{EAF1FB}83.09 & 80.06 \\
\bottomrule
\end{tabular*}
\end{minipage}\par\medskip

\par\smallskip\noindent\begin{minipage}{\linewidth}
\centering
\setlength{\abovecaptionskip}{3pt}
\setlength{\belowcaptionskip}{3pt}
\captionof{table}{Domain-level T for translations into Japanese, averaged over available source languages.}
\label{tab:target_domain_ja}
\scriptsize
\setlength{\tabcolsep}{1.5pt}
\renewcommand{\arraystretch}{1.12}
\begin{tabular*}{\linewidth}{@{\extracolsep{\fill}}l*{10}{r}@{}}
\toprule
\textbf{Model} & \textbf{Transport} & \textbf{Dining} & \textbf{Products} & \textbf{Tourism} & \textbf{\shortstack{Health\\care}} & \textbf{Safety} & \textbf{\shortstack{Social\\Media}} & \textbf{\shortstack{Publi-\\cations}} & \textbf{Finance} & \textbf{\shortstack{Public\\Services}} \\
\midrule
\multicolumn{11}{l}{\textbf{Closed-weight models}} \\
GPT-5.6 Luna & 55.09 & 71.89 & 52.86 & 53.73 & 69.54 & 57.69 & 64.59 & 60.96 & 59.78 & 58.80 \\
Claude Sonnet 5 & 67.25 & 76.49 & 69.70 & 69.54 & 75.52 & 77.56 & 69.45 & 73.16 & 73.53 & 73.76 \\
Muse Spark 1.3 & 69.30 & 74.77 & 71.80 & 69.89 & 76.75 & 75.52 & 71.80 & 78.12 & 77.71 & 72.08 \\
Qwen3.8 Max & 68.85 & 75.25 & \cellcolor[HTML]{EAF1FB}80.05 & 71.42 & 78.23 & \cellcolor[HTML]{EAF1FB}82.11 & 73.18 & \cellcolor[HTML]{EAF1FB}81.40 & \cellcolor[HTML]{EAF1FB}84.17 & 77.47 \\
Gemini 3.8 Flash & 69.92 & 73.14 & 72.34 & 75.80 & 79.13 & 80.72 & 71.46 & 76.64 & 78.01 & 74.87 \\
Doubao Seed 2.1 Pro & 71.22 & 77.18 & 72.03 & 71.43 & 77.49 & 79.89 & 70.85 & 76.92 & 79.13 & 75.94 \\
\bottomrule
\end{tabular*}
\end{minipage}\par\medskip

\par\smallskip\noindent\begin{minipage}{\linewidth}
\centering
\setlength{\abovecaptionskip}{3pt}
\setlength{\belowcaptionskip}{3pt}
\captionof{table}{Domain-level T for translations into Korean, averaged over available source languages.}
\label{tab:target_domain_ko}
\scriptsize
\setlength{\tabcolsep}{1.5pt}
\renewcommand{\arraystretch}{1.12}
\begin{tabular*}{\linewidth}{@{\extracolsep{\fill}}l*{10}{r}@{}}
\toprule
\textbf{Model} & \textbf{Transport} & \textbf{Dining} & \textbf{Products} & \textbf{Tourism} & \textbf{\shortstack{Health\\care}} & \textbf{Safety} & \textbf{\shortstack{Social\\Media}} & \textbf{\shortstack{Publi-\\cations}} & \textbf{Finance} & \textbf{\shortstack{Public\\Services}} \\
\midrule
\multicolumn{11}{l}{\textbf{Closed-weight models}} \\
GPT-5.6 Luna & 52.59 & 60.55 & 53.10 & 53.14 & 62.94 & 60.83 & 57.85 & 57.44 & 49.98 & 66.25 \\
Claude Sonnet 5 & 66.30 & \cellcolor[HTML]{EAF1FB}77.85 & 67.88 & 70.04 & 74.13 & 83.75 & 67.74 & 72.53 & 71.24 & 80.76 \\
Muse Spark 1.3 & 62.99 & 71.06 & 73.24 & 72.31 & 72.03 & 76.10 & 71.67 & 78.63 & 65.54 & 75.46 \\
Qwen3.8 Max & 67.83 & 70.18 & 73.07 & 70.33 & 72.88 & \cellcolor[HTML]{EAF1FB}84.41 & 75.60 & 76.92 & 67.01 & 77.25 \\
Gemini 3.8 Flash & 70.37 & 77.34 & 77.53 & \cellcolor[HTML]{EAF1FB}78.38 & 74.21 & 81.42 & 67.57 & 80.98 & 77.32 & 80.40 \\
Doubao Seed 2.1 Pro & 65.67 & 72.51 & 75.30 & 75.59 & \cellcolor[HTML]{EAF1FB}79.24 & 79.91 & 71.96 & 79.31 & \cellcolor[HTML]{EAF1FB}79.09 & \cellcolor[HTML]{EAF1FB}84.45 \\
\bottomrule
\end{tabular*}
\end{minipage}\par\medskip

\par\smallskip\noindent\begin{minipage}{\linewidth}
\centering
\setlength{\abovecaptionskip}{3pt}
\setlength{\belowcaptionskip}{3pt}
\captionof{table}{Domain-level T for translations into Chinese, averaged over available source languages.}
\label{tab:target_domain_zh}
\scriptsize
\setlength{\tabcolsep}{1.5pt}
\renewcommand{\arraystretch}{1.12}
\begin{tabular*}{\linewidth}{@{\extracolsep{\fill}}l*{10}{r}@{}}
\toprule
\textbf{Model} & \textbf{Transport} & \textbf{Dining} & \textbf{Products} & \textbf{Tourism} & \textbf{\shortstack{Health\\care}} & \textbf{Safety} & \textbf{\shortstack{Social\\Media}} & \textbf{\shortstack{Publi-\\cations}} & \textbf{Finance} & \textbf{\shortstack{Public\\Services}} \\
\midrule
\multicolumn{11}{l}{\textbf{Closed-weight models}} \\
GPT-5.6 Luna & 49.37 & 56.22 & 53.68 & 58.68 & 67.49 & 60.49 & 55.79 & 58.46 & 58.75 & 58.38 \\
Claude Sonnet 5 & 65.09 & 71.39 & 57.18 & 70.30 & 69.78 & 73.55 & 67.91 & 68.84 & 75.54 & 78.87 \\
Muse Spark 1.3 & 63.50 & \cellcolor[HTML]{EAF1FB}74.83 & 67.63 & 72.17 & 75.90 & 76.83 & 73.36 & 76.09 & 75.93 & 76.73 \\
Qwen3.8 Max & 63.24 & 70.70 & 69.89 & 73.30 & 74.88 & 80.43 & 76.25 & 74.63 & 75.96 & 71.67 \\
Gemini 3.8 Flash & 64.06 & 72.17 & 73.37 & 76.63 & \cellcolor[HTML]{EAF1FB}77.34 & 75.43 & 70.11 & 74.29 & 74.91 & 76.04 \\
Doubao Seed 2.1 Pro & 67.95 & 73.33 & 72.59 & 76.56 & 75.79 & 79.47 & 80.20 & 78.18 & \cellcolor[HTML]{EAF1FB}80.03 & 74.63 \\
\bottomrule
\end{tabular*}
\end{minipage}\par\medskip

\paragraph{Target-dependent domain rankings.}
GPT-6 Astra leads in 116 of the 220 target-language--domain combinations, followed by Doubao Seed 2.1 Pro in 46 and Gemini 3.8 Flash in 37.
The benchmark-wide ranking therefore does not hold uniformly for every translation target.
For Products translated into English, Gemini scores 76.21 compared with GPT-6 Astra's 74.65; for Chinese, GPT-6 Astra scores 77.82 compared with Gemini's 73.37 (Tables~\ref{tab:target_domain_en} and~\ref{tab:target_domain_zh}).

The strongest model also varies across domains for a fixed target.
For Japanese, Qwen3.8 Max leads Products at 80.05 and Finance at 84.17, while GPT-6 Astra leads Healthcare at 83.76 (Table~\ref{tab:target_domain_ja}).
These target-conditioned profiles complement the source-language analysis by describing performance for the language in which the translation is delivered.

\clearpage
\subsection{Complete Language-Pair Results}
\label{sec:full_direction_matrix}

The following tables show the complete source-to-target T matrix for each image-input model on a 0--100 scale.
Rows denote source languages and columns denote target languages; same-language diagonal cells are omitted.

\par\smallskip\noindent\begin{minipage}{\linewidth}
\centering
\captionof{table}{Direction-level translation quality for Gemma-4 E4B. Rows are source languages and columns are target languages.}
\label{tab:direction_gemma_4_e4b}
\vspace{3pt}
\scriptsize
\setlength{\tabcolsep}{1.4pt}
\renewcommand{\arraystretch}{1.0}
\begin{tabular*}{\linewidth}{@{\extracolsep{\fill}}l*{22}{r}@{}}
\toprule
\textbf{Src$\backslash$Tgt} & \textbf{en} & \textbf{cs} & \textbf{de} & \textbf{nl} & \textbf{fr} & \textbf{it} & \textbf{es} & \textbf{pt} & \textbf{tr} & \textbf{id} & \textbf{ms} & \textbf{vi} & \textbf{ru} & \textbf{uk} & \textbf{ar} & \textbf{he} & \textbf{hi} & \textbf{my} & \textbf{th} & \textbf{ja} & \textbf{ko} & \textbf{zh} \\
\midrule
\textbf{en} & -- & 18.4 & 26.2 & 23.3 & 15.8 & 19.2 & 23.0 & 19.8 & 18.4 & 22.7 & 19.3 & 19.9 & 22.6 & 26.4 & 23.1 & 17.1 & 18.8 & 27.6 & 23.1 & 24.2 & 20.6 & 21.3 \\
\textbf{cs} & 22.3 & -- & 17.0 & 15.4 & 27.3 & 23.2 & 19.2 & 18.9 & 17.6 & 19.4 & 21.6 & 21.5 & 27.3 & 30.4 & 22.7 & 18.4 & 24.5 & 27.0 & 27.9 & 16.2 & 14.5 & 16.0 \\
\textbf{de} & 14.4 & 13.5 & -- & 19.0 & 17.1 & 19.7 & 17.1 & 16.3 & 15.2 & 21.7 & 15.8 & 15.7 & 22.7 & 16.6 & 16.9 & 20.3 & 18.6 & 17.6 & 27.2 & 17.4 & 13.5 & 16.4 \\
\textbf{nl} & 23.0 & 22.3 & 15.8 & -- & 20.0 & 16.8 & 11.1 & 16.3 & 21.5 & 19.3 & 16.3 & 9.7 & 20.1 & 11.2 & 12.2 & 22.5 & 18.1 & 22.7 & 26.5 & 16.5 & 15.5 & 18.2 \\
\textbf{fr} & 31.5 & 20.5 & 22.0 & 32.3 & -- & 28.3 & 25.6 & 22.1 & 30.9 & 26.2 & 26.9 & 30.5 & 26.4 & 26.5 & 20.9 & 23.5 & 28.5 & 28.5 & 32.0 & 23.4 & 23.7 & 24.8 \\
\textbf{it} & 19.2 & 16.2 & 16.6 & 20.1 & 20.9 & -- & 14.6 & 19.0 & 17.6 & 19.4 & 22.3 & 22.6 & 21.0 & 16.0 & 19.4 & 18.1 & 16.0 & 17.0 & 21.7 & 20.9 & 22.8 & 17.6 \\
\textbf{es} & 23.1 & 25.5 & 20.9 & 25.3 & 29.6 & 25.1 & -- & 23.0 & 18.0 & 22.2 & 22.1 & 19.6 & 19.0 & 21.4 & 27.6 & 28.4 & 18.3 & 26.3 & 25.5 & 18.6 & 23.5 & 21.0 \\
\bottomrule
\end{tabular*}
\end{minipage}\par\medskip

\par\smallskip\noindent\begin{minipage}{\linewidth}
\centering
\captionof{table}{Direction-level translation quality for GLM-4.1V 9B. Rows are source languages and columns are target languages.}
\label{tab:direction_glm_4_1v_9b}
\vspace{3pt}
\scriptsize
\setlength{\tabcolsep}{1.4pt}
\renewcommand{\arraystretch}{1.0}
\begin{tabular*}{\linewidth}{@{\extracolsep{\fill}}l*{22}{r}@{}}
\toprule
\textbf{Src$\backslash$Tgt} & \textbf{en} & \textbf{cs} & \textbf{de} & \textbf{nl} & \textbf{fr} & \textbf{it} & \textbf{es} & \textbf{pt} & \textbf{tr} & \textbf{id} & \textbf{ms} & \textbf{vi} & \textbf{ru} & \textbf{uk} & \textbf{ar} & \textbf{he} & \textbf{hi} & \textbf{my} & \textbf{th} & \textbf{ja} & \textbf{ko} & \textbf{zh} \\
\midrule
\textbf{en} & -- & 40.7 & 34.2 & 38.2 & 35.5 & 41.0 & 35.8 & 31.5 & 37.2 & 31.3 & 33.0 & 32.1 & 36.7 & 40.4 & 30.3 & 36.9 & 37.5 & 30.5 & 35.1 & 35.9 & 42.9 & 33.6 \\
\textbf{cs} & 32.0 & -- & 30.5 & 29.9 & 33.1 & 30.1 & 28.4 & 37.7 & 30.3 & 21.6 & 32.0 & 27.3 & 36.4 & 40.2 & 36.3 & 37.2 & 38.5 & 40.3 & 38.1 & 26.1 & 26.6 & 38.7 \\
\textbf{de} & 35.8 & 33.6 & -- & 47.3 & 41.1 & 35.3 & 39.0 & 38.4 & 32.8 & 35.8 & 35.6 & 32.0 & 27.8 & 39.1 & 31.3 & 40.1 & 40.2 & 37.2 & 40.9 & 36.5 & 35.6 & 34.7 \\
\textbf{nl} & 58.4 & 54.0 & 48.6 & -- & 41.6 & 51.8 & 39.2 & 51.3 & 39.2 & 43.2 & 36.4 & 41.0 & 40.0 & 46.3 & 28.6 & 35.9 & 49.2 & 52.7 & 47.6 & 43.0 & 43.0 & 45.2 \\
\textbf{fr} & 44.0 & 36.9 & 43.4 & 43.1 & -- & 40.8 & 42.4 & 51.0 & 39.3 & 42.3 & 44.7 & 33.5 & 49.6 & 38.1 & 40.1 & 41.7 & 52.8 & 47.4 & 43.9 & 51.7 & 47.1 & 37.9 \\
\textbf{it} & 32.5 & 34.3 & 35.8 & 32.8 & 35.7 & -- & 38.9 & 41.2 & 31.8 & 37.1 & 34.6 & 36.7 & 34.5 & 28.1 & 32.3 & 33.7 & 30.2 & 36.3 & 36.3 & 33.5 & 34.0 & 35.1 \\
\textbf{es} & 42.2 & 43.9 & 37.5 & 43.3 & 38.6 & 41.5 & -- & 44.1 & 39.9 & 31.4 & 35.0 & 38.1 & 34.2 & 43.4 & 31.7 & 33.0 & 33.1 & 40.7 & 47.2 & 32.9 & 36.5 & 40.4 \\
\bottomrule
\end{tabular*}
\end{minipage}\par\medskip

\par\smallskip\noindent\begin{minipage}{\linewidth}
\centering
\captionof{table}{Direction-level translation quality for GLM-5.3 Flash. Rows are source languages and columns are target languages.}
\label{tab:direction_glm_5_3_flash}
\vspace{3pt}
\scriptsize
\setlength{\tabcolsep}{1.4pt}
\renewcommand{\arraystretch}{1.0}
\begin{tabular*}{\linewidth}{@{\extracolsep{\fill}}l*{22}{r}@{}}
\toprule
\textbf{Src$\backslash$Tgt} & \textbf{en} & \textbf{cs} & \textbf{de} & \textbf{nl} & \textbf{fr} & \textbf{it} & \textbf{es} & \textbf{pt} & \textbf{tr} & \textbf{id} & \textbf{ms} & \textbf{vi} & \textbf{ru} & \textbf{uk} & \textbf{ar} & \textbf{he} & \textbf{hi} & \textbf{my} & \textbf{th} & \textbf{ja} & \textbf{ko} & \textbf{zh} \\
\midrule
\textbf{en} & -- & 49.4 & 58.4 & 55.8 & 49.9 & 49.9 & 55.8 & 53.1 & 56.3 & 55.8 & 51.4 & 51.1 & 59.4 & 55.0 & 62.0 & 54.8 & 57.5 & 61.6 & 67.8 & 57.8 & 63.2 & 55.1 \\
\textbf{cs} & 61.1 & -- & 58.7 & 69.1 & 64.0 & 59.8 & 56.9 & 66.4 & 71.1 & 61.7 & 71.1 & 59.0 & 68.4 & 73.4 & 71.1 & 61.5 & 66.9 & 75.5 & 61.8 & 61.8 & 65.0 & 65.3 \\
\textbf{de} & 39.0 & 50.9 & -- & 55.8 & 54.5 & 54.5 & 52.3 & 49.8 & 47.7 & 64.2 & 51.8 & 54.2 & 52.9 & 57.4 & 49.0 & 53.0 & 55.6 & 52.6 & 60.6 & 51.0 & 50.7 & 56.2 \\
\textbf{nl} & 75.1 & 83.5 & 77.9 & -- & 75.9 & 82.5 & 68.4 & 78.9 & 82.6 & 76.7 & 85.3 & 71.9 & 77.8 & 79.0 & 77.8 & 76.2 & 79.5 & 82.1 & 78.0 & 80.6 & 78.3 & 74.5 \\
\textbf{fr} & 60.3 & 68.7 & 64.3 & 57.9 & -- & 57.0 & 66.2 & 60.4 & 67.5 & 59.1 & 65.0 & 70.0 & 61.3 & 69.0 & 62.4 & 69.8 & 76.5 & 66.3 & 64.3 & 67.2 & 69.7 & 65.3 \\
\textbf{it} & 46.1 & 57.5 & 57.9 & 48.4 & 53.4 & -- & 48.0 & 52.4 & 50.3 & 54.8 & 54.3 & 54.6 & 57.5 & 50.3 & 56.7 & 55.8 & 54.2 & 65.0 & 54.7 & 54.0 & 50.7 & 52.2 \\
\textbf{es} & 61.3 & 60.7 & 58.9 & 58.3 & 58.7 & 65.4 & -- & 59.7 & 61.8 & 57.0 & 60.9 & 56.9 & 54.6 & 57.9 & 71.8 & 61.3 & 56.6 & 66.1 & 62.2 & 61.7 & 59.3 & 60.5 \\
\bottomrule
\end{tabular*}
\end{minipage}\par\medskip

\par\smallskip\noindent\begin{minipage}{\linewidth}
\centering
\captionof{table}{Direction-level translation quality for GPT-5.6 Luna. Rows are source languages and columns are target languages.}
\label{tab:direction_gpt_5_6_luna}
\vspace{3pt}
\scriptsize
\setlength{\tabcolsep}{1.4pt}
\renewcommand{\arraystretch}{1.0}
\begin{tabular*}{\linewidth}{@{\extracolsep{\fill}}l*{22}{r}@{}}
\toprule
\textbf{Src$\backslash$Tgt} & \textbf{en} & \textbf{cs} & \textbf{de} & \textbf{nl} & \textbf{fr} & \textbf{it} & \textbf{es} & \textbf{pt} & \textbf{tr} & \textbf{id} & \textbf{ms} & \textbf{vi} & \textbf{ru} & \textbf{uk} & \textbf{ar} & \textbf{he} & \textbf{hi} & \textbf{my} & \textbf{th} & \textbf{ja} & \textbf{ko} & \textbf{zh} \\
\midrule
\textbf{en} & -- & 58.2 & 52.0 & 54.6 & 48.1 & 53.5 & 52.4 & 50.8 & 42.4 & 53.0 & 58.1 & 45.8 & 48.3 & 56.1 & 61.8 & 59.1 & 47.8 & 64.6 & 61.2 & 52.2 & 53.5 & 55.0 \\
\textbf{cs} & 66.5 & -- & 60.0 & 71.5 & 72.4 & 62.2 & 65.2 & 66.8 & 74.0 & 61.9 & 61.9 & 69.6 & 62.5 & 68.7 & 68.2 & 70.3 & 69.2 & 83.8 & 71.5 & 68.7 & 65.3 & 67.1 \\
\textbf{de} & 45.7 & 47.5 & -- & 60.3 & 53.6 & 51.5 & 58.7 & 47.9 & 52.9 & 59.6 & 53.7 & 52.3 & 53.5 & 63.0 & 52.4 & 49.4 & 54.0 & 57.4 & 57.1 & 49.3 & 49.6 & 49.5 \\
\textbf{nl} & 57.8 & 72.2 & 68.5 & -- & 70.7 & 64.5 & 63.1 & 56.0 & 69.6 & 73.8 & 68.7 & 70.5 & 72.8 & 63.1 & 63.3 & 75.5 & 69.8 & 71.6 & 64.8 & 66.9 & 46.2 & 64.7 \\
\textbf{fr} & 62.5 & 67.9 & 67.5 & 65.9 & -- & 58.8 & 63.8 & 68.3 & 65.7 & 69.2 & 71.0 & 67.0 & 61.9 & 66.7 & 69.8 & 73.4 & 68.4 & 73.3 & 67.8 & 62.3 & 62.4 & 64.6 \\
\textbf{it} & 55.9 & 61.0 & 60.8 & 54.7 & 59.0 & -- & 56.9 & 54.8 & 59.9 & 62.7 & 57.1 & 59.7 & 59.2 & 55.8 & 53.9 & 59.1 & 56.2 & 65.1 & 64.4 & 62.4 & 58.2 & 62.3 \\
\textbf{es} & 66.8 & 61.8 & 65.2 & 68.0 & 58.4 & 71.7 & -- & 64.3 & 65.7 & 62.7 & 62.4 & 71.3 & 65.0 & 72.4 & 67.3 & 73.7 & 72.1 & 67.0 & 62.5 & 72.6 & 67.4 & 72.5 \\
\bottomrule
\end{tabular*}
\end{minipage}\par\medskip

\par\smallskip\noindent\begin{minipage}{\linewidth}
\centering
\captionof{table}{Direction-level translation quality for DeepSeek Flash. Rows are source languages and columns are target languages.}
\label{tab:direction_deepseek_flash}
\vspace{3pt}
\scriptsize
\setlength{\tabcolsep}{1.4pt}
\renewcommand{\arraystretch}{1.0}
\begin{tabular*}{\linewidth}{@{\extracolsep{\fill}}l*{22}{r}@{}}
\toprule
\textbf{Src$\backslash$Tgt} & \textbf{en} & \textbf{cs} & \textbf{de} & \textbf{nl} & \textbf{fr} & \textbf{it} & \textbf{es} & \textbf{pt} & \textbf{tr} & \textbf{id} & \textbf{ms} & \textbf{vi} & \textbf{ru} & \textbf{uk} & \textbf{ar} & \textbf{he} & \textbf{hi} & \textbf{my} & \textbf{th} & \textbf{ja} & \textbf{ko} & \textbf{zh} \\
\midrule
\textbf{en} & -- & 52.8 & 50.7 & 55.9 & 55.2 & 57.1 & 47.5 & 50.8 & 48.6 & 48.6 & 57.9 & 54.5 & 59.1 & 53.2 & 61.7 & 64.0 & 55.8 & 59.4 & 57.4 & 46.2 & 48.1 & 54.1 \\
\textbf{cs} & 72.6 & -- & 69.6 & 78.1 & 81.1 & 80.3 & 70.6 & 84.3 & 76.0 & 73.7 & 75.7 & 73.1 & 78.5 & 77.7 & 80.9 & 73.2 & 85.1 & 79.6 & 76.8 & 71.7 & 75.8 & 76.6 \\
\textbf{de} & 46.6 & 57.6 & -- & 58.7 & 56.4 & 54.4 & 60.7 & 50.3 & 57.1 & 66.9 & 58.5 & 52.8 & 60.7 & 61.8 & 59.1 & 58.0 & 54.3 & 61.5 & 68.3 & 54.3 & 61.7 & 58.6 \\
\textbf{nl} & 71.5 & 70.1 & 67.0 & -- & 65.0 & 88.8 & 67.5 & 69.7 & 72.6 & 75.7 & 69.2 & 63.5 & 79.5 & 75.2 & 66.1 & 71.7 & 83.4 & 84.2 & 78.7 & 76.7 & 68.2 & 72.8 \\
\textbf{fr} & 68.6 & 70.1 & 63.8 & 72.4 & -- & 65.5 & 61.0 & 64.0 & 71.7 & 65.0 & 71.0 & 68.5 & 64.0 & 75.0 & 67.1 & 72.5 & 70.7 & 76.8 & 72.4 & 64.0 & 66.6 & 63.9 \\
\textbf{it} & 52.8 & 61.1 & 52.8 & 57.2 & 58.9 & -- & 56.5 & 44.9 & 55.7 & 60.7 & 55.0 & 57.4 & 57.5 & 46.7 & 47.6 & 52.9 & 50.5 & 62.7 & 55.5 & 58.5 & 49.7 & 51.9 \\
\textbf{es} & 64.1 & 74.2 & 70.7 & 67.8 & 69.7 & 70.1 & -- & 73.7 & 68.0 & 64.7 & 66.7 & 71.1 & 68.2 & 75.3 & 74.4 & 69.0 & 68.7 & 77.6 & 73.0 & 70.9 & 68.9 & 72.5 \\
\bottomrule
\end{tabular*}
\end{minipage}\par\medskip

\par\smallskip\noindent\begin{minipage}{\linewidth}
\centering
\captionof{table}{Direction-level translation quality for Qwen3.8 27B. Rows are source languages and columns are target languages.}
\label{tab:direction_qwen3_8_27b_fp8}
\vspace{3pt}
\scriptsize
\setlength{\tabcolsep}{1.4pt}
\renewcommand{\arraystretch}{1.0}
\begin{tabular*}{\linewidth}{@{\extracolsep{\fill}}l*{22}{r}@{}}
\toprule
\textbf{Src$\backslash$Tgt} & \textbf{en} & \textbf{cs} & \textbf{de} & \textbf{nl} & \textbf{fr} & \textbf{it} & \textbf{es} & \textbf{pt} & \textbf{tr} & \textbf{id} & \textbf{ms} & \textbf{vi} & \textbf{ru} & \textbf{uk} & \textbf{ar} & \textbf{he} & \textbf{hi} & \textbf{my} & \textbf{th} & \textbf{ja} & \textbf{ko} & \textbf{zh} \\
\midrule
\textbf{en} & -- & 63.1 & 60.9 & 56.9 & 57.8 & 65.2 & 55.0 & 57.8 & 52.1 & 66.8 & 62.6 & 61.7 & 60.5 & 64.8 & 62.0 & 65.3 & 59.3 & 62.1 & 62.2 & 67.8 & 60.8 & 60.4 \\
\textbf{cs} & 67.7 & -- & 61.8 & 64.6 & 64.4 & 65.8 & 61.9 & 55.5 & 58.9 & 65.5 & 69.8 & 60.3 & 61.5 & 65.9 & 73.0 & 61.0 & 68.3 & 71.5 & 61.7 & 68.7 & 54.4 & 63.7 \\
\textbf{de} & 56.3 & 53.6 & -- & 63.0 & 56.4 & 55.0 & 59.9 & 60.7 & 61.5 & 65.6 & 57.2 & 57.0 & 56.0 & 62.0 & 59.5 & 53.3 & 64.8 & 62.2 & 63.5 & 57.7 & 54.7 & 59.3 \\
\textbf{nl} & 67.2 & 68.2 & 59.0 & -- & 67.3 & 64.8 & 77.3 & 68.1 & 70.1 & 68.5 & 71.2 & 70.7 & 72.9 & 67.8 & 65.3 & 78.3 & 74.8 & 79.4 & 84.2 & 85.1 & 73.4 & 74.7 \\
\textbf{fr} & 62.9 & 67.4 & 63.5 & 57.0 & -- & 55.9 & 61.6 & 52.3 & 65.1 & 64.7 & 71.4 & 61.0 & 65.2 & 60.7 & 66.7 & 60.4 & 67.0 & 73.8 & 73.2 & 62.8 & 62.3 & 60.6 \\
\textbf{it} & 59.3 & 60.5 & 62.9 & 65.2 & 62.6 & -- & 57.5 & 59.9 & 69.1 & 57.9 & 58.6 & 59.4 & 65.1 & 53.3 & 56.8 & 58.1 & 56.4 & 70.4 & 62.3 & 62.4 & 56.5 & 55.5 \\
\textbf{es} & 65.4 & 63.3 & 69.0 & 60.0 & 65.4 & 69.9 & -- & 63.3 & 64.0 & 63.1 & 64.9 & 66.3 & 69.3 & 67.8 & 70.6 & 58.5 & 75.1 & 68.6 & 67.8 & 71.9 & 67.0 & 68.5 \\
\bottomrule
\end{tabular*}
\end{minipage}\par\medskip

\par\smallskip\noindent\begin{minipage}{\linewidth}
\centering
\captionof{table}{Direction-level translation quality for Claude Sonnet 5. Rows are source languages and columns are target languages.}
\label{tab:direction_claude_sonnet_5}
\vspace{3pt}
\scriptsize
\setlength{\tabcolsep}{1.4pt}
\renewcommand{\arraystretch}{1.0}
\begin{tabular*}{\linewidth}{@{\extracolsep{\fill}}l*{22}{r}@{}}
\toprule
\textbf{Src$\backslash$Tgt} & \textbf{en} & \textbf{cs} & \textbf{de} & \textbf{nl} & \textbf{fr} & \textbf{it} & \textbf{es} & \textbf{pt} & \textbf{tr} & \textbf{id} & \textbf{ms} & \textbf{vi} & \textbf{ru} & \textbf{uk} & \textbf{ar} & \textbf{he} & \textbf{hi} & \textbf{my} & \textbf{th} & \textbf{ja} & \textbf{ko} & \textbf{zh} \\
\midrule
\textbf{en} & -- & 65.4 & 61.3 & 57.6 & 54.5 & 61.9 & 52.7 & 57.9 & 56.3 & 56.5 & 60.7 & 68.8 & 63.9 & 68.0 & 63.6 & 64.0 & 58.2 & 72.3 & 63.8 & 51.8 & 58.4 & 58.5 \\
\textbf{cs} & 83.6 & -- & 84.3 & 80.1 & 74.0 & 81.2 & 78.5 & 82.5 & 86.8 & 76.6 & 85.1 & 75.7 & 81.4 & 80.9 & 83.0 & 80.5 & 78.2 & 83.4 & 81.8 & 85.1 & 79.8 & 78.4 \\
\textbf{de} & 57.1 & 65.3 & -- & 61.1 & 65.7 & 63.7 & 61.3 & 57.2 & 61.0 & 67.9 & 59.7 & 62.1 & 64.1 & 69.3 & 55.6 & 65.1 & 68.8 & 76.0 & 62.1 & 65.0 & 62.8 & 54.9 \\
\textbf{nl} & 79.4 & 84.9 & 84.2 & -- & 93.0 & 94.0 & 84.3 & 79.9 & 92.7 & 87.7 & 90.5 & 92.7 & 91.2 & 88.9 & 85.1 & 90.9 & 89.5 & 85.9 & 87.5 & 86.1 & 82.6 & 83.4 \\
\textbf{fr} & 69.3 & 76.5 & 75.4 & 66.7 & -- & 67.7 & 76.3 & 74.6 & 69.3 & 67.0 & 81.3 & 74.2 & 67.6 & 67.4 & 75.2 & 69.9 & 79.8 & 79.4 & 77.8 & 75.3 & 75.1 & 64.6 \\
\textbf{it} & 63.2 & 67.5 & 63.7 & 69.1 & 68.3 & -- & 67.5 & 58.8 & 60.5 & 68.8 & 70.3 & 68.1 & 66.8 & 65.9 & 65.4 & 67.2 & 66.8 & 75.5 & 69.2 & 68.3 & 71.2 & 60.0 \\
\textbf{es} & 66.9 & 64.5 & 64.6 & 66.4 & 60.7 & 71.5 & -- & 66.8 & 65.4 & 67.5 & 67.8 & 68.7 & 68.7 & 69.6 & 76.0 & 71.9 & 72.9 & 73.8 & 71.7 & 60.2 & 66.1 & 70.7 \\
\bottomrule
\end{tabular*}
\end{minipage}\par\medskip

\par\smallskip\noindent\begin{minipage}{\linewidth}
\centering
\captionof{table}{Direction-level translation quality for Muse Spark 1.3. Rows are source languages and columns are target languages.}
\label{tab:direction_muse_spark_1_3}
\vspace{3pt}
\scriptsize
\setlength{\tabcolsep}{1.4pt}
\renewcommand{\arraystretch}{1.0}
\begin{tabular*}{\linewidth}{@{\extracolsep{\fill}}l*{22}{r}@{}}
\toprule
\textbf{Src$\backslash$Tgt} & \textbf{en} & \textbf{cs} & \textbf{de} & \textbf{nl} & \textbf{fr} & \textbf{it} & \textbf{es} & \textbf{pt} & \textbf{tr} & \textbf{id} & \textbf{ms} & \textbf{vi} & \textbf{ru} & \textbf{uk} & \textbf{ar} & \textbf{he} & \textbf{hi} & \textbf{my} & \textbf{th} & \textbf{ja} & \textbf{ko} & \textbf{zh} \\
\midrule
\textbf{en} & -- & 64.5 & 59.5 & 61.0 & 61.8 & 65.5 & 50.0 & 62.4 & 63.7 & 63.5 & 64.2 & 58.3 & 65.1 & 70.6 & 68.6 & 68.4 & 66.4 & 75.3 & 68.6 & 60.2 & 64.2 & 58.9 \\
\textbf{cs} & 82.7 & -- & 80.7 & 81.2 & 88.3 & 80.8 & 85.4 & 89.8 & 95.9 & 81.7 & 89.0 & 82.4 & 87.2 & 94.8 & 83.5 & 83.2 & 86.4 & 84.3 & 88.4 & 79.8 & 75.5 & 87.5 \\
\textbf{de} & 46.3 & 61.3 & -- & 62.4 & 64.4 & 62.5 & 56.2 & 58.1 & 67.2 & 67.4 & 60.2 & 62.2 & 62.9 & 70.4 & 63.0 & 63.1 & 61.6 & 70.2 & 63.9 & 56.0 & 58.9 & 62.9 \\
\textbf{nl} & 81.4 & 79.8 & 75.1 & -- & 83.3 & 81.9 & 86.6 & 85.0 & 85.3 & 77.4 & 80.8 & 84.9 & 82.5 & 77.3 & 87.5 & 85.9 & 76.3 & 86.1 & 79.1 & 83.9 & 81.8 & 83.7 \\
\textbf{fr} & 64.4 & 86.3 & 73.1 & 75.7 & -- & 76.7 & 78.0 & 71.9 & 81.9 & 76.5 & 78.1 & 64.8 & 82.5 & 79.4 & 77.1 & 77.4 & 81.7 & 83.4 & 79.7 & 76.5 & 72.0 & 71.7 \\
\textbf{it} & 60.9 & 72.1 & 68.1 & 70.3 & 68.4 & -- & 65.8 & 60.9 & 64.0 & 66.9 & 68.6 & 65.8 & 63.8 & 70.7 & 69.6 & 71.3 & 64.8 & 67.0 & 67.2 & 76.2 & 62.6 & 64.2 \\
\textbf{es} & 69.0 & 80.2 & 75.8 & 73.5 & 79.5 & 71.6 & -- & 73.4 & 75.1 & 77.1 & 74.2 & 72.8 & 71.9 & 75.8 & 67.7 & 73.0 & 71.3 & 76.9 & 71.1 & 71.1 & 73.3 & 76.6 \\
\bottomrule
\end{tabular*}
\end{minipage}\par\medskip

\par\smallskip\noindent\begin{minipage}{\linewidth}
\centering
\captionof{table}{Direction-level translation quality for Qwen3.8 Max. Rows are source languages and columns are target languages.}
\label{tab:direction_qwen3_8_max_0902}
\vspace{3pt}
\scriptsize
\setlength{\tabcolsep}{1.4pt}
\renewcommand{\arraystretch}{1.0}
\begin{tabular*}{\linewidth}{@{\extracolsep{\fill}}l*{22}{r}@{}}
\toprule
\textbf{Src$\backslash$Tgt} & \textbf{en} & \textbf{cs} & \textbf{de} & \textbf{nl} & \textbf{fr} & \textbf{it} & \textbf{es} & \textbf{pt} & \textbf{tr} & \textbf{id} & \textbf{ms} & \textbf{vi} & \textbf{ru} & \textbf{uk} & \textbf{ar} & \textbf{he} & \textbf{hi} & \textbf{my} & \textbf{th} & \textbf{ja} & \textbf{ko} & \textbf{zh} \\
\midrule
\textbf{en} & -- & 74.9 & 69.4 & 62.2 & 64.5 & 70.2 & 61.6 & 60.7 & 64.7 & 66.0 & 73.0 & 67.4 & 73.3 & 63.9 & 69.3 & 76.2 & 68.2 & 63.5 & 73.3 & 79.6 & 64.7 & 65.6 \\
\textbf{cs} & 85.4 & -- & 76.7 & 80.2 & 85.2 & 82.6 & 77.2 & 88.6 & 86.1 & 83.0 & 91.9 & 85.2 & 85.5 & 87.0 & 89.8 & 88.5 & 87.2 & 75.2 & 86.9 & 82.6 & 80.9 & 88.9 \\
\textbf{de} & 60.8 & 67.9 & -- & 72.3 & 77.2 & 68.9 & 63.5 & 65.0 & 69.5 & 72.0 & 71.2 & 70.3 & 65.2 & 75.1 & 72.0 & 63.7 & 69.9 & 68.3 & 70.6 & 70.3 & 70.2 & 70.4 \\
\textbf{nl} & 83.2 & 83.7 & 80.5 & -- & 83.6 & 77.0 & 88.2 & 81.8 & 80.5 & 80.7 & 85.1 & 85.1 & 89.8 & 82.6 & 80.3 & 82.9 & 85.0 & 70.3 & 78.7 & 86.3 & 78.4 & 77.9 \\
\textbf{fr} & 75.1 & 77.2 & 74.7 & 74.5 & -- & 65.8 & 74.0 & 70.4 & 79.9 & 72.9 & 78.2 & 74.9 & 72.8 & 75.1 & 77.2 & 78.7 & 79.6 & 64.3 & 74.9 & 75.4 & 73.1 & 66.7 \\
\textbf{it} & 72.6 & 72.3 & 74.2 & 75.4 & 69.0 & -- & 69.0 & 65.4 & 76.4 & 70.5 & 68.7 & 73.4 & 72.3 & 74.2 & 72.8 & 73.3 & 70.1 & 66.0 & 78.5 & 77.0 & 71.7 & 68.4 \\
\textbf{es} & 74.8 & 78.4 & 74.4 & 77.8 & 71.4 & 78.7 & -- & 69.9 & 76.8 & 72.2 & 78.1 & 77.5 & 76.0 & 77.4 & 83.3 & 77.4 & 71.2 & 68.0 & 76.8 & 77.0 & 76.1 & 75.1 \\
\bottomrule
\end{tabular*}
\end{minipage}\par\medskip

\par\smallskip\noindent\begin{minipage}{\linewidth}
\centering
\captionof{table}{Direction-level translation quality for Gemini 3.8 Flash. Rows are source languages and columns are target languages.}
\label{tab:direction_gemini_3_8_flash}
\vspace{3pt}
\scriptsize
\setlength{\tabcolsep}{1.4pt}
\renewcommand{\arraystretch}{1.0}
\begin{tabular*}{\linewidth}{@{\extracolsep{\fill}}l*{22}{r}@{}}
\toprule
\textbf{Src$\backslash$Tgt} & \textbf{en} & \textbf{cs} & \textbf{de} & \textbf{nl} & \textbf{fr} & \textbf{it} & \textbf{es} & \textbf{pt} & \textbf{tr} & \textbf{id} & \textbf{ms} & \textbf{vi} & \textbf{ru} & \textbf{uk} & \textbf{ar} & \textbf{he} & \textbf{hi} & \textbf{my} & \textbf{th} & \textbf{ja} & \textbf{ko} & \textbf{zh} \\
\midrule
\textbf{en} & -- & 69.4 & 60.0 & 68.1 & 67.0 & 71.1 & 57.5 & 65.0 & 61.8 & 66.1 & 67.9 & 65.8 & 64.0 & 67.4 & 64.7 & 78.4 & 61.5 & 73.7 & 74.8 & 67.7 & 68.6 & 63.3 \\
\textbf{cs} & 88.1 & -- & 88.6 & 84.0 & 83.7 & 81.5 & 90.0 & 86.5 & 87.0 & 87.0 & 85.3 & 83.3 & 87.4 & 86.7 & 82.0 & 89.8 & 89.0 & 88.8 & 84.8 & 85.5 & 86.7 & 84.8 \\
\textbf{de} & 65.5 & 71.9 & -- & 75.2 & 77.2 & 66.3 & 68.8 & 69.7 & 66.3 & 73.9 & 68.0 & 64.2 & 67.0 & 76.7 & 64.4 & 70.0 & 71.8 & 78.9 & 71.8 & 72.5 & 66.8 & 68.7 \\
\textbf{nl} & 88.6 & 93.8 & 93.5 & -- & 93.5 & 89.4 & 87.0 & 87.8 & 86.7 & 97.8 & 93.8 & 87.9 & 98.0 & 89.7 & 88.2 & 83.3 & 85.8 & 86.0 & 81.6 & 80.4 & 94.4 & 82.2 \\
\textbf{fr} & 73.7 & 80.9 & 79.6 & 75.6 & -- & 70.8 & 79.7 & 73.7 & 79.4 & 73.9 & 79.3 & 83.1 & 72.8 & 77.7 & 84.7 & 83.6 & 80.1 & 83.0 & 81.3 & 76.2 & 77.4 & 74.9 \\
\textbf{it} & 68.3 & 74.5 & 70.8 & 76.8 & 72.4 & -- & 69.5 & 73.9 & 69.1 & 72.9 & 69.8 & 73.5 & 69.1 & 76.1 & 69.3 & 70.7 & 71.6 & 76.5 & 72.4 & 74.8 & 73.0 & 67.9 \\
\textbf{es} & 66.9 & 83.5 & 69.5 & 73.3 & 67.1 & 72.3 & -- & 74.9 & 77.2 & 74.5 & 78.9 & 83.2 & 73.0 & 71.8 & 79.6 & 79.3 & 75.8 & 79.0 & 75.6 & 76.1 & 73.8 & 70.3 \\
\bottomrule
\end{tabular*}
\end{minipage}\par\medskip

\par\smallskip\noindent\begin{minipage}{\linewidth}
\centering
\captionof{table}{Direction-level translation quality for Doubao Seed 2.1 Pro. Rows are source languages and columns are target languages.}
\label{tab:direction_doubao_seed_2_1_pro_260915}
\vspace{3pt}
\scriptsize
\setlength{\tabcolsep}{1.4pt}
\renewcommand{\arraystretch}{1.0}
\begin{tabular*}{\linewidth}{@{\extracolsep{\fill}}l*{22}{r}@{}}
\toprule
\textbf{Src$\backslash$Tgt} & \textbf{en} & \textbf{cs} & \textbf{de} & \textbf{nl} & \textbf{fr} & \textbf{it} & \textbf{es} & \textbf{pt} & \textbf{tr} & \textbf{id} & \textbf{ms} & \textbf{vi} & \textbf{ru} & \textbf{uk} & \textbf{ar} & \textbf{he} & \textbf{hi} & \textbf{my} & \textbf{th} & \textbf{ja} & \textbf{ko} & \textbf{zh} \\
\midrule
\textbf{en} & -- & 74.0 & 70.7 & 66.7 & 73.8 & 79.9 & 63.8 & 76.8 & 72.9 & 73.8 & 75.8 & 73.0 & 73.3 & 70.9 & 77.9 & 74.8 & 61.4 & 67.5 & 75.2 & 69.4 & 68.5 & 74.4 \\
\textbf{cs} & 90.9 & -- & 82.8 & 85.6 & 92.3 & 84.0 & 90.3 & 81.8 & 92.2 & 89.1 & 86.5 & 85.5 & 94.2 & 88.4 & 84.4 & 81.7 & 83.9 & 92.9 & 88.2 & 86.7 & 79.4 & 91.4 \\
\textbf{de} & 64.9 & 74.1 & -- & 67.9 & 68.8 & 68.3 & 66.0 & 62.7 & 64.3 & 74.2 & 70.6 & 61.6 & 77.1 & 76.5 & 65.9 & 74.6 & 74.2 & 71.1 & 73.6 & 64.6 & 67.1 & 70.8 \\
\textbf{nl} & 82.3 & 86.4 & 84.8 & -- & 92.6 & 86.3 & 75.8 & 90.6 & 90.7 & 91.4 & 78.9 & 90.6 & 86.7 & 90.9 & 89.2 & 81.6 & 88.4 & 76.4 & 86.3 & 86.0 & 88.5 & 83.6 \\
\textbf{fr} & 75.3 & 88.2 & 75.0 & 80.8 & -- & 80.9 & 81.8 & 80.2 & 73.6 & 78.9 & 79.4 & 79.9 & 82.5 & 75.8 & 84.1 & 82.7 & 82.7 & 91.0 & 81.7 & 73.4 & 85.8 & 77.3 \\
\textbf{it} & 71.3 & 73.6 & 76.4 & 76.2 & 68.9 & -- & 68.0 & 62.5 & 73.8 & 73.1 & 64.5 & 72.6 & 68.1 & 68.0 & 73.5 & 72.8 & 68.1 & 70.2 & 75.6 & 71.2 & 74.3 & 66.9 \\
\textbf{es} & 74.6 & 86.1 & 76.1 & 78.1 & 77.3 & 79.2 & -- & 74.2 & 82.8 & 81.1 & 81.9 & 77.8 & 82.0 & 85.0 & 86.0 & 70.9 & 77.6 & 78.6 & 77.0 & 79.5 & 78.5 & 76.1 \\
\bottomrule
\end{tabular*}
\end{minipage}\par\medskip

\par\smallskip\noindent\begin{minipage}{\linewidth}
\centering
\captionof{table}{Direction-level translation quality for GPT-6 Astra. Rows are source languages and columns are target languages.}
\label{tab:direction_gpt_6_astra}
\vspace{3pt}
\scriptsize
\setlength{\tabcolsep}{1.4pt}
\renewcommand{\arraystretch}{1.0}
\begin{tabular*}{\linewidth}{@{\extracolsep{\fill}}l*{22}{r}@{}}
\toprule
\textbf{Src$\backslash$Tgt} & \textbf{en} & \textbf{cs} & \textbf{de} & \textbf{nl} & \textbf{fr} & \textbf{it} & \textbf{es} & \textbf{pt} & \textbf{tr} & \textbf{id} & \textbf{ms} & \textbf{vi} & \textbf{ru} & \textbf{uk} & \textbf{ar} & \textbf{he} & \textbf{hi} & \textbf{my} & \textbf{th} & \textbf{ja} & \textbf{ko} & \textbf{zh} \\
\midrule
\textbf{en} & -- & 81.9 & 79.3 & 78.5 & 70.3 & 77.3 & 82.6 & 76.4 & 83.6 & 76.9 & 78.1 & 80.1 & 79.6 & 77.9 & 79.3 & 79.7 & 75.1 & 84.7 & 82.3 & 78.4 & 76.7 & 78.5 \\
\textbf{cs} & 90.9 & -- & 82.7 & 92.0 & 92.1 & 86.5 & 91.5 & 89.5 & 87.8 & 92.0 & 87.8 & 87.6 & 83.4 & 94.0 & 87.6 & 90.6 & 88.0 & 89.1 & 87.4 & 84.6 & 93.1 & 88.3 \\
\textbf{de} & 73.9 & 75.0 & -- & 75.9 & 73.1 & 68.8 & 74.2 & 62.2 & 79.9 & 76.8 & 72.1 & 72.4 & 77.2 & 83.2 & 73.3 & 69.3 & 72.6 & 85.1 & 75.3 & 78.9 & 71.2 & 74.1 \\
\textbf{nl} & 86.4 & 83.9 & 88.6 & -- & 85.8 & 87.0 & 79.6 & 91.8 & 89.9 & 90.7 & 84.2 & 86.2 & 84.3 & 80.8 & 88.5 & 85.5 & 85.8 & 90.7 & 92.1 & 86.5 & 81.9 & 89.4 \\
\textbf{fr} & 76.6 & 85.6 & 79.0 & 80.1 & -- & 74.2 & 79.0 & 88.8 & 88.7 & 80.2 & 82.8 & 82.9 & 80.1 & 80.0 & 82.0 & 84.4 & 84.5 & 85.9 & 85.1 & 82.5 & 77.7 & 80.0 \\
\textbf{it} & 69.1 & 76.0 & 77.2 & 79.4 & 79.6 & -- & 76.7 & 69.1 & 74.3 & 74.5 & 76.6 & 76.4 & 78.7 & 82.5 & 73.8 & 76.0 & 78.1 & 79.7 & 76.1 & 79.9 & 74.5 & 76.2 \\
\textbf{es} & 78.0 & 84.1 & 81.0 & 82.0 & 77.7 & 78.6 & -- & 80.8 & 78.6 & 80.5 & 84.2 & 86.1 & 82.1 & 81.9 & 88.3 & 91.1 & 77.5 & 87.9 & 77.8 & 79.2 & 82.7 & 79.9 \\
\bottomrule
\end{tabular*}
\end{minipage}\par\medskip

\clearpage
\subsection{Dimension Profiles by Source Language}
\label{app:source_dimension_results}

Tables~\ref{tab:source_dimensions_1}--\ref{tab:source_dimensions_6} report T, K, and V separately for each model and source-image language, averaged over valid target languages.
Each table presents two models.
Blue and purple cells mark the best scores across the full closed- and open-weight groups, respectively, for the same language and metric.

\par\smallskip\noindent\begin{minipage}{\linewidth}
\centering
\setlength{\abovecaptionskip}{3pt}
\setlength{\belowcaptionskip}{3pt}
\captionof{table}{Source-language T, K, and V for GPT-5.6 Luna and Claude Sonnet 5.}
\label{tab:source_dimensions_1}
\scriptsize
\setlength{\tabcolsep}{1.5pt}
\renewcommand{\arraystretch}{1.12}
\begin{tabular*}{\linewidth}{@{\extracolsep{\fill}}l*{6}{r}@{}}
\toprule
 & \multicolumn{3}{c}{\textbf{GPT-5.6 Luna}} & \multicolumn{3}{c}{\textbf{Claude Sonnet 5}} \\
\cmidrule(lr){2-4}\cmidrule(lr){5-7}
\textbf{Source language} & T & K & V & T & K & V \\
\midrule
English & 53.73 & 82.94 & 59.80 & 60.77 & 86.78 & 57.40 \\
Czech & 67.96 & 76.06 & 74.83 & 80.99 & 87.00 & 78.48 \\
German & 53.29 & 89.32 & 58.94 & 63.14 & 84.02 & 65.22 \\
Dutch & 66.38 & 72.53 & 71.70 & 87.33 & 87.40 & 76.93 \\
French & 66.59 & 89.26 & 65.32 & 72.89 & 90.41 & 58.43 \\
Italian & 59.00 & 73.63 & 54.02 & 66.78 & 81.52 & 56.65 \\
Spanish & 67.18 & 92.56 & 65.11 & 68.21 & 93.63 & 63.90 \\
Portuguese & 63.21 & 96.71 & 71.70 & 66.11 & 97.47 & 69.36 \\
Turkish & 61.18 & 92.20 & 59.26 & 67.06 & 83.24 & 51.73 \\
Indonesian & 60.93 & 94.16 & 72.42 & 77.24 & 95.67 & 77.27 \\
Malay & 65.21 & 98.55 & 75.75 & 72.59 & 98.56 & 77.22 \\
Vietnamese & 65.20 & 91.20 & 70.91 & 75.10 & 89.88 & 71.86 \\
Russian & 59.33 & 82.57 & 61.42 & 75.23 & 86.07 & 67.87 \\
Ukrainian & 50.22 & 90.45 & 48.96 & 71.12 & 98.21 & 75.95 \\
Arabic & 49.50 & 81.80 & 56.18 & 66.71 & 80.35 & 71.70 \\
Hebrew & 37.87 & 58.73 & 42.21 & 67.46 & 68.21 & 72.77 \\
Hindi & 52.52 & 53.86 & 59.04 & 76.58 & \cellcolor[HTML]{EAF1FB}93.06 & 73.25 \\
Burmese & 29.00 & 33.90 & 37.28 & 43.75 & 39.77 & 53.53 \\
Thai & 62.73 & 83.82 & 69.59 & 68.57 & 83.70 & 72.66 \\
Japanese & 56.27 & 63.32 & 48.71 & 66.85 & 81.09 & 57.03 \\
Korean & 56.50 & 56.99 & 52.28 & 72.74 & 76.21 & 71.08 \\
Chinese & 41.58 & 61.38 & 51.17 & 65.73 & 79.00 & 68.19 \\
\bottomrule
\end{tabular*}
\end{minipage}\par\medskip

\par\smallskip\noindent\begin{minipage}{\linewidth}
\centering
\setlength{\abovecaptionskip}{3pt}
\setlength{\belowcaptionskip}{3pt}
\captionof{table}{Source-language T, K, and V for Muse Spark 1.3 and Qwen3.8 Max.}
\label{tab:source_dimensions_2}
\scriptsize
\setlength{\tabcolsep}{1.5pt}
\renewcommand{\arraystretch}{1.12}
\begin{tabular*}{\linewidth}{@{\extracolsep{\fill}}l*{6}{r}@{}}
\toprule
 & \multicolumn{3}{c}{\textbf{Muse Spark 1.3}} & \multicolumn{3}{c}{\textbf{Qwen3.8 Max}} \\
\cmidrule(lr){2-4}\cmidrule(lr){5-7}
\textbf{Source language} & T & K & V & T & K & V \\
\midrule
English & 63.84 & 81.83 & 64.79 & 68.20 & 84.55 & 74.87 \\
Czech & 85.17 & 94.60 & 85.71 & 84.50 & 92.85 & 87.79 \\
German & 61.96 & 84.86 & 61.95 & 69.25 & 93.89 & 75.31 \\
Dutch & 82.18 & 86.92 & 81.34 & 81.98 & 85.79 & 80.35 \\
French & 76.60 & \cellcolor[HTML]{EAF1FB}94.72 & 66.25 & 74.07 & 93.83 & 73.39 \\
Italian & 67.10 & 80.29 & 60.20 & 71.95 & \cellcolor[HTML]{EAF1FB}96.26 & 71.67 \\
Spanish & 73.86 & 97.94 & 68.03 & 75.62 & 92.88 & 72.47 \\
Portuguese & 67.58 & 95.69 & 82.53 & 73.83 & 96.49 & 86.19 \\
Turkish & 73.23 & 86.46 & 66.37 & 70.32 & 93.67 & 60.90 \\
Indonesian & 76.25 & 95.05 & 81.41 & 76.63 & 88.59 & 78.22 \\
Malay & 74.48 & \cellcolor[HTML]{EAF1FB}99.12 & 87.54 & 72.85 & 96.27 & 78.79 \\
Vietnamese & 81.63 & 97.05 & 78.57 & 79.74 & 94.03 & 78.33 \\
Russian & \cellcolor[HTML]{EAF1FB}79.25 & \cellcolor[HTML]{EAF1FB}86.74 & 74.69 & 77.08 & 85.10 & 73.40 \\
Ukrainian & 70.51 & 98.50 & 71.20 & 69.16 & 99.75 & 72.90 \\
Arabic & 70.18 & 89.71 & 77.36 & 68.70 & 84.95 & 77.96 \\
Hebrew & 72.73 & 77.10 & 77.23 & 67.64 & 69.36 & 70.32 \\
Hindi & 78.39 & 91.26 & 75.54 & 79.28 & 88.19 & 81.18 \\
Burmese & 43.40 & 39.87 & 54.86 & 42.19 & 39.68 & 57.47 \\
Thai & 72.30 & 86.77 & 78.08 & 70.16 & 86.22 & 78.73 \\
Japanese & 69.14 & 88.56 & 61.35 & 72.74 & 84.80 & 64.71 \\
Korean & 74.50 & 87.95 & 73.53 & 80.93 & 92.42 & 81.70 \\
Chinese & 66.09 & 72.64 & 65.69 & 72.71 & 82.98 & 78.02 \\
\bottomrule
\end{tabular*}
\end{minipage}\par\medskip

\par\smallskip\noindent\begin{minipage}{\linewidth}
\centering
\setlength{\abovecaptionskip}{3pt}
\setlength{\belowcaptionskip}{3pt}
\captionof{table}{Source-language T, K, and V for Gemini 3.8 Flash and Doubao Seed 2.1 Pro.}
\label{tab:source_dimensions_3}
\scriptsize
\setlength{\tabcolsep}{1.5pt}
\renewcommand{\arraystretch}{1.12}
\begin{tabular*}{\linewidth}{@{\extracolsep{\fill}}l*{6}{r}@{}}
\toprule
 & \multicolumn{3}{c}{\textbf{Gemini 3.8 Flash}} & \multicolumn{3}{c}{\textbf{Doubao Seed 2.1 Pro}} \\
\cmidrule(lr){2-4}\cmidrule(lr){5-7}
\textbf{Source language} & T & K & V & T & K & V \\
\midrule
English & 66.86 & 83.81 & 70.67 & 72.12 & 88.55 & 81.98 \\
Czech & 86.21 & 96.00 & 86.54 & 87.25 & 96.30 & 89.02 \\
German & 70.28 & 86.24 & 75.82 & 69.47 & \cellcolor[HTML]{EAF1FB}96.84 & 83.53 \\
Dutch & \cellcolor[HTML]{EAF1FB}89.02 & 85.83 & 78.72 & 86.09 & \cellcolor[HTML]{EAF1FB}92.94 & \cellcolor[HTML]{EAF1FB}83.57 \\
French & 78.16 & 90.46 & 75.94 & 80.53 & 93.75 & 82.36 \\
Italian & 72.04 & 93.67 & 71.29 & 70.93 & 80.14 & 76.02 \\
Spanish & 75.03 & 96.51 & 68.66 & 79.07 & 96.62 & 80.97 \\
Portuguese & 75.06 & \cellcolor[HTML]{EAF1FB}98.13 & 84.75 & 69.24 & 97.34 & 80.50 \\
Turkish & 76.43 & \cellcolor[HTML]{EAF1FB}94.95 & 71.74 & 77.59 & 94.87 & \cellcolor[HTML]{EAF1FB}81.79 \\
Indonesian & \cellcolor[HTML]{EAF1FB}83.26 & \cellcolor[HTML]{EAF1FB}97.27 & 81.84 & 80.08 & 94.12 & 91.42 \\
Malay & 76.07 & 97.70 & 86.58 & 80.78 & 98.91 & \cellcolor[HTML]{EAF1FB}93.08 \\
Vietnamese & 82.22 & 96.77 & 78.56 & 81.89 & 97.95 & 85.02 \\
Russian & 75.59 & 85.58 & 70.80 & 76.90 & 86.31 & 76.98 \\
Ukrainian & 69.57 & \cellcolor[HTML]{EAF1FB}100.00 & 72.51 & 70.36 & 99.35 & 73.80 \\
Arabic & 70.41 & 87.65 & 74.93 & \cellcolor[HTML]{EAF1FB}71.14 & \cellcolor[HTML]{EAF1FB}92.23 & \cellcolor[HTML]{EAF1FB}79.64 \\
Hebrew & 71.42 & \cellcolor[HTML]{EAF1FB}77.88 & 77.68 & 67.48 & 69.79 & 77.40 \\
Hindi & 80.20 & 91.30 & 75.26 & 82.32 & 91.64 & \cellcolor[HTML]{EAF1FB}84.90 \\
Burmese & 42.83 & 39.66 & 54.40 & 45.98 & 41.13 & 63.92 \\
Thai & 69.74 & 87.18 & 79.16 & 71.51 & 86.87 & 84.21 \\
Japanese & 72.85 & \cellcolor[HTML]{EAF1FB}90.00 & 66.13 & 74.86 & 88.02 & 73.41 \\
Korean & 81.80 & \cellcolor[HTML]{EAF1FB}96.00 & 82.70 & \cellcolor[HTML]{EAF1FB}82.24 & 94.14 & \cellcolor[HTML]{EAF1FB}85.86 \\
Chinese & 72.22 & 81.48 & 73.55 & \cellcolor[HTML]{EAF1FB}75.15 & \cellcolor[HTML]{EAF1FB}90.33 & \cellcolor[HTML]{EAF1FB}81.85 \\
\bottomrule
\end{tabular*}
\end{minipage}\par\medskip

\par\smallskip\noindent\begin{minipage}{\linewidth}
\centering
\setlength{\abovecaptionskip}{3pt}
\setlength{\belowcaptionskip}{3pt}
\captionof{table}{Source-language T, K, and V for GPT-6 Astra and Gemma-4 E4B.}
\label{tab:source_dimensions_4}
\scriptsize
\setlength{\tabcolsep}{1.5pt}
\renewcommand{\arraystretch}{1.12}
\begin{tabular*}{\linewidth}{@{\extracolsep{\fill}}l*{6}{r}@{}}
\toprule
 & \multicolumn{3}{c}{\textbf{GPT-6 Astra}} & \multicolumn{3}{c}{\textbf{Gemma-4 E4B}} \\
\cmidrule(lr){2-4}\cmidrule(lr){5-7}
\textbf{Source language} & T & K & V & T & K & V \\
\midrule
English & \cellcolor[HTML]{EAF1FB}78.92 & \cellcolor[HTML]{EAF1FB}88.64 & \cellcolor[HTML]{EAF1FB}91.33 & 21.47 & 41.16 & 16.88 \\
Czech & \cellcolor[HTML]{EAF1FB}88.88 & \cellcolor[HTML]{EAF1FB}98.60 & \cellcolor[HTML]{EAF1FB}91.76 & 21.34 & 29.24 & 15.02 \\
German & \cellcolor[HTML]{EAF1FB}74.48 & 93.65 & \cellcolor[HTML]{EAF1FB}87.49 & 17.74 & 27.28 & 11.96 \\
Dutch & 86.64 & 82.22 & 80.39 & 17.90 & 48.85 & 24.21 \\
French & \cellcolor[HTML]{EAF1FB}81.91 & 92.90 & \cellcolor[HTML]{EAF1FB}86.47 & 26.43 & 28.76 & 23.43 \\
Italian & \cellcolor[HTML]{EAF1FB}76.41 & 95.90 & \cellcolor[HTML]{EAF1FB}81.22 & 18.99 & 29.29 & 14.33 \\
Spanish & \cellcolor[HTML]{EAF1FB}81.89 & \cellcolor[HTML]{EAF1FB}99.76 & \cellcolor[HTML]{EAF1FB}81.98 & 23.05 & 34.70 & 18.97 \\
Portuguese & \cellcolor[HTML]{EAF1FB}76.10 & 97.50 & \cellcolor[HTML]{EAF1FB}90.58 & 20.46 & 53.06 & 18.12 \\
Turkish & \cellcolor[HTML]{EAF1FB}79.04 & 94.39 & 79.25 & 28.63 & 46.62 & 22.42 \\
Indonesian & 79.44 & 93.08 & \cellcolor[HTML]{EAF1FB}91.59 & 25.55 & 33.52 & 26.16 \\
Malay & \cellcolor[HTML]{EAF1FB}81.00 & 96.90 & 87.61 & 25.61 & 54.98 & 20.62 \\
Vietnamese & \cellcolor[HTML]{EAF1FB}85.07 & \cellcolor[HTML]{EAF1FB}98.40 & \cellcolor[HTML]{EAF1FB}88.57 & 17.79 & 29.27 & 12.84 \\
Russian & 79.14 & 86.47 & \cellcolor[HTML]{EAF1FB}84.72 & 19.01 & 28.73 & 16.38 \\
Ukrainian & \cellcolor[HTML]{EAF1FB}73.00 & 99.14 & \cellcolor[HTML]{EAF1FB}79.58 & 22.54 & 52.75 & 24.84 \\
Arabic & 70.87 & 86.23 & 79.15 & 21.45 & 32.34 & 22.18 \\
Hebrew & \cellcolor[HTML]{EAF1FB}74.68 & 77.85 & \cellcolor[HTML]{EAF1FB}82.25 & 20.05 & 26.89 & 19.51 \\
Hindi & \cellcolor[HTML]{EAF1FB}83.41 & 89.57 & 80.85 & 36.94 & 51.81 & 35.66 \\
Burmese & \cellcolor[HTML]{EAF1FB}51.22 & \cellcolor[HTML]{EAF1FB}51.29 & \cellcolor[HTML]{EAF1FB}66.22 & 8.36 & 5.83 & 9.16 \\
Thai & \cellcolor[HTML]{EAF1FB}76.42 & \cellcolor[HTML]{EAF1FB}88.01 & \cellcolor[HTML]{EAF1FB}86.97 & 15.50 & 15.87 & 15.43 \\
Japanese & \cellcolor[HTML]{EAF1FB}79.23 & 87.74 & \cellcolor[HTML]{EAF1FB}82.01 & 10.28 & 15.53 & 9.44 \\
Korean & 80.18 & 86.62 & 85.16 & 13.40 & 16.06 & 11.37 \\
Chinese & 73.88 & 86.48 & 77.48 & 9.15 & 17.45 & 10.54 \\
\bottomrule
\end{tabular*}
\end{minipage}\par\medskip

\par\smallskip\noindent\begin{minipage}{\linewidth}
\centering
\setlength{\abovecaptionskip}{3pt}
\setlength{\belowcaptionskip}{3pt}
\captionof{table}{Source-language T, K, and V for GLM-4.1V 9B and GLM-5.3 Flash.}
\label{tab:source_dimensions_5}
\scriptsize
\setlength{\tabcolsep}{1.5pt}
\renewcommand{\arraystretch}{1.12}
\begin{tabular*}{\linewidth}{@{\extracolsep{\fill}}l*{6}{r}@{}}
\toprule
 & \multicolumn{3}{c}{\textbf{GLM-4.1V 9B}} & \multicolumn{3}{c}{\textbf{GLM-5.3 Flash}} \\
\cmidrule(lr){2-4}\cmidrule(lr){5-7}
\textbf{Source language} & T & K & V & T & K & V \\
\midrule
English & 35.74 & 63.93 & 39.42 & 56.24 & 82.04 & 59.21 \\
Czech & 32.92 & 40.99 & 47.09 & 65.22 & 70.29 & 70.43 \\
German & 36.66 & 58.27 & 40.70 & 53.03 & 84.82 & 57.28 \\
Dutch & 44.58 & 78.07 & 58.60 & \cellcolor[HTML]{F0EBFA}78.22 & \cellcolor[HTML]{F0EBFA}96.82 & 66.36 \\
French & 43.43 & 74.67 & 39.38 & 65.15 & 88.42 & 56.96 \\
Italian & 34.54 & 45.59 & 39.07 & 53.76 & \cellcolor[HTML]{F0EBFA}77.10 & 57.17 \\
Spanish & 38.50 & 62.46 & 42.42 & 60.56 & 91.95 & 52.87 \\
Portuguese & 38.54 & 85.60 & 49.51 & \cellcolor[HTML]{F0EBFA}63.83 & \cellcolor[HTML]{F0EBFA}95.48 & 66.09 \\
Turkish & 36.86 & 65.43 & 36.84 & 58.73 & 86.11 & 48.76 \\
Indonesian & 37.73 & 76.34 & 55.94 & 66.33 & \cellcolor[HTML]{F0EBFA}93.63 & 69.00 \\
Malay & 43.70 & 77.27 & 52.60 & \cellcolor[HTML]{F0EBFA}64.05 & 93.59 & 66.90 \\
Vietnamese & 29.55 & 50.00 & 35.32 & 60.63 & 84.28 & 64.58 \\
Russian & 38.21 & 51.06 & 37.83 & 64.50 & \cellcolor[HTML]{F0EBFA}84.04 & 62.00 \\
Ukrainian & 24.23 & 63.07 & 37.52 & 59.79 & 95.14 & 63.09 \\
Arabic & 14.60 & 14.44 & 14.41 & 38.56 & 55.65 & 49.31 \\
Hebrew & 6.25 & 9.82 & 8.89 & 35.11 & 51.89 & 46.18 \\
Hindi & 9.30 & 8.39 & 12.51 & 48.61 & 74.04 & 55.57 \\
Burmese & 15.13 & 12.20 & 16.60 & 22.90 & 12.33 & 30.78 \\
Thai & 11.67 & 14.35 & 14.17 & 33.46 & 41.51 & 43.39 \\
Japanese & 32.46 & 38.33 & 31.79 & 58.31 & \cellcolor[HTML]{F0EBFA}73.49 & 52.95 \\
Korean & 15.76 & 20.43 & 20.23 & 48.84 & 49.55 & 52.53 \\
Chinese & 33.79 & 45.55 & 38.34 & 58.82 & 75.22 & 62.23 \\
\bottomrule
\end{tabular*}
\end{minipage}\par\medskip

\par\smallskip\noindent\begin{minipage}{\linewidth}
\centering
\setlength{\abovecaptionskip}{3pt}
\setlength{\belowcaptionskip}{3pt}
\captionof{table}{Source-language T, K, and V for DeepSeek Flash and Qwen3.8 27B.}
\label{tab:source_dimensions_6}
\scriptsize
\setlength{\tabcolsep}{1.5pt}
\renewcommand{\arraystretch}{1.12}
\begin{tabular*}{\linewidth}{@{\extracolsep{\fill}}l*{6}{r}@{}}
\toprule
 & \multicolumn{3}{c}{\textbf{DeepSeek Flash}} & \multicolumn{3}{c}{\textbf{Qwen3.8 27B}} \\
\cmidrule(lr){2-4}\cmidrule(lr){5-7}
\textbf{Source language} & T & K & V & T & K & V \\
\midrule
English & 54.21 & \cellcolor[HTML]{F0EBFA}83.91 & \cellcolor[HTML]{F0EBFA}68.58 & \cellcolor[HTML]{F0EBFA}61.20 & 82.60 & 65.76 \\
Czech & \cellcolor[HTML]{F0EBFA}76.71 & \cellcolor[HTML]{F0EBFA}85.25 & \cellcolor[HTML]{F0EBFA}83.57 & 64.10 & 81.65 & 80.63 \\
German & 58.01 & 89.89 & 64.80 & \cellcolor[HTML]{F0EBFA}59.01 & \cellcolor[HTML]{F0EBFA}92.96 & \cellcolor[HTML]{F0EBFA}68.48 \\
Dutch & 73.19 & 85.29 & 78.24 & 71.82 & 80.92 & \cellcolor[HTML]{F0EBFA}80.94 \\
French & \cellcolor[HTML]{F0EBFA}68.31 & 87.55 & \cellcolor[HTML]{F0EBFA}65.85 & 63.60 & \cellcolor[HTML]{F0EBFA}89.95 & 58.89 \\
Italian & 54.60 & 54.23 & 53.98 & \cellcolor[HTML]{F0EBFA}60.46 & 68.06 & \cellcolor[HTML]{F0EBFA}65.50 \\
Spanish & \cellcolor[HTML]{F0EBFA}70.44 & 89.31 & \cellcolor[HTML]{F0EBFA}68.11 & 66.65 & \cellcolor[HTML]{F0EBFA}92.29 & 67.08 \\
Portuguese & 61.05 & 93.09 & 77.37 & 63.51 & 95.16 & \cellcolor[HTML]{F0EBFA}78.14 \\
Turkish & \cellcolor[HTML]{F0EBFA}64.43 & 82.75 & \cellcolor[HTML]{F0EBFA}59.61 & 61.99 & \cellcolor[HTML]{F0EBFA}92.00 & 58.03 \\
Indonesian & \cellcolor[HTML]{F0EBFA}73.50 & 92.14 & \cellcolor[HTML]{F0EBFA}80.25 & 65.32 & 85.66 & 71.57 \\
Malay & 62.97 & \cellcolor[HTML]{F0EBFA}97.91 & \cellcolor[HTML]{F0EBFA}80.72 & 63.22 & 90.97 & 78.34 \\
Vietnamese & \cellcolor[HTML]{F0EBFA}65.84 & \cellcolor[HTML]{F0EBFA}91.91 & \cellcolor[HTML]{F0EBFA}71.71 & 62.71 & 84.50 & 66.45 \\
Russian & 64.35 & 80.09 & 69.00 & \cellcolor[HTML]{F0EBFA}71.99 & 82.65 & \cellcolor[HTML]{F0EBFA}71.13 \\
Ukrainian & 58.88 & \cellcolor[HTML]{F0EBFA}95.77 & \cellcolor[HTML]{F0EBFA}63.14 & \cellcolor[HTML]{F0EBFA}62.27 & 92.14 & 62.14 \\
Arabic & 52.24 & 72.85 & 62.38 & \cellcolor[HTML]{F0EBFA}58.45 & \cellcolor[HTML]{F0EBFA}80.55 & \cellcolor[HTML]{F0EBFA}68.26 \\
Hebrew & 49.34 & \cellcolor[HTML]{F0EBFA}60.34 & 53.88 & \cellcolor[HTML]{F0EBFA}54.13 & 46.26 & \cellcolor[HTML]{F0EBFA}66.75 \\
Hindi & \cellcolor[HTML]{F0EBFA}65.89 & 75.21 & \cellcolor[HTML]{F0EBFA}69.40 & 60.37 & \cellcolor[HTML]{F0EBFA}83.16 & 67.27 \\
Burmese & \cellcolor[HTML]{F0EBFA}34.74 & \cellcolor[HTML]{F0EBFA}35.57 & \cellcolor[HTML]{F0EBFA}47.97 & 25.35 & 20.88 & 36.73 \\
Thai & \cellcolor[HTML]{F0EBFA}53.94 & 63.49 & \cellcolor[HTML]{F0EBFA}68.64 & 52.42 & \cellcolor[HTML]{F0EBFA}67.88 & 66.22 \\
Japanese & 59.58 & 69.00 & 52.96 & \cellcolor[HTML]{F0EBFA}64.44 & 73.32 & \cellcolor[HTML]{F0EBFA}65.25 \\
Korean & 59.34 & 61.79 & 59.71 & \cellcolor[HTML]{F0EBFA}63.30 & \cellcolor[HTML]{F0EBFA}71.43 & \cellcolor[HTML]{F0EBFA}68.96 \\
Chinese & 61.78 & \cellcolor[HTML]{F0EBFA}77.20 & 68.30 & \cellcolor[HTML]{F0EBFA}64.79 & 72.82 & \cellcolor[HTML]{F0EBFA}75.06 \\
\bottomrule
\end{tabular*}
\end{minipage}\par\medskip

\paragraph{Alignment and variation across languages.}
Across the 12 models within each source language, Pearson correlations range from 0.9277 to 0.9917 for T--V and from 0.8635 to 0.9932 for T--K, showing strong alignment.
GPT-6 Astra leads T on 16 of the 22 source languages, V on 15, and K on six.
Within GPT-6 Astra, correlations across the 22 source-language means are 0.6820 for T--V and 0.7274 for T--K, lower than the cross-model correlations in Section~\ref{sec:dimension_analysis}.

Burmese has the lowest T for 11 of the 12 models, the lowest K for 11, and the lowest V for ten.
For GPT-6 Astra, its Burmese scores are 51.22 in T, 51.29 in K, and 66.22 in V, compared with 78.92, 88.64, and 91.33 on English.
The lower scores identify Burmese as a shared difficulty under the three rubric diagnostics; they do not isolate the source of the errors or exclude annotation and judge limitations.

\end{document}